\documentclass[runningheads]{llncs}
 
\usepackage{eccv}

\usepackage{eccvabbrv}

\usepackage{graphicx}
\usepackage{booktabs}
\usepackage[dvipsnames]{xcolor}
\usepackage[accsupp]{axessibility}  

\usepackage{hyperref}

\usepackage{orcidlink}

\begin{document}

\title{Learning visual representations for compositional analysis of artworks and photographs}
\titlerunning{Learning visual representations for compositional analysis}

\author{Fatemeh Behrad\orcidlink{0000-0003-2629-0854} \and
Tinne Tuytelaars
\orcidlink{0000-0003-3307-9723} \and
Johan Wagemans\orcidlink{0000-0002-7970-1541}}

\authorrunning{F. Behrad et al.}

\institute{
KU Leuven University\\ Belgium
}

\maketitle
\vspace{-8mm}
\begin{figure}
    \centering
    \captionsetup{type=figure}
    \includegraphics[width=0.70\linewidth]{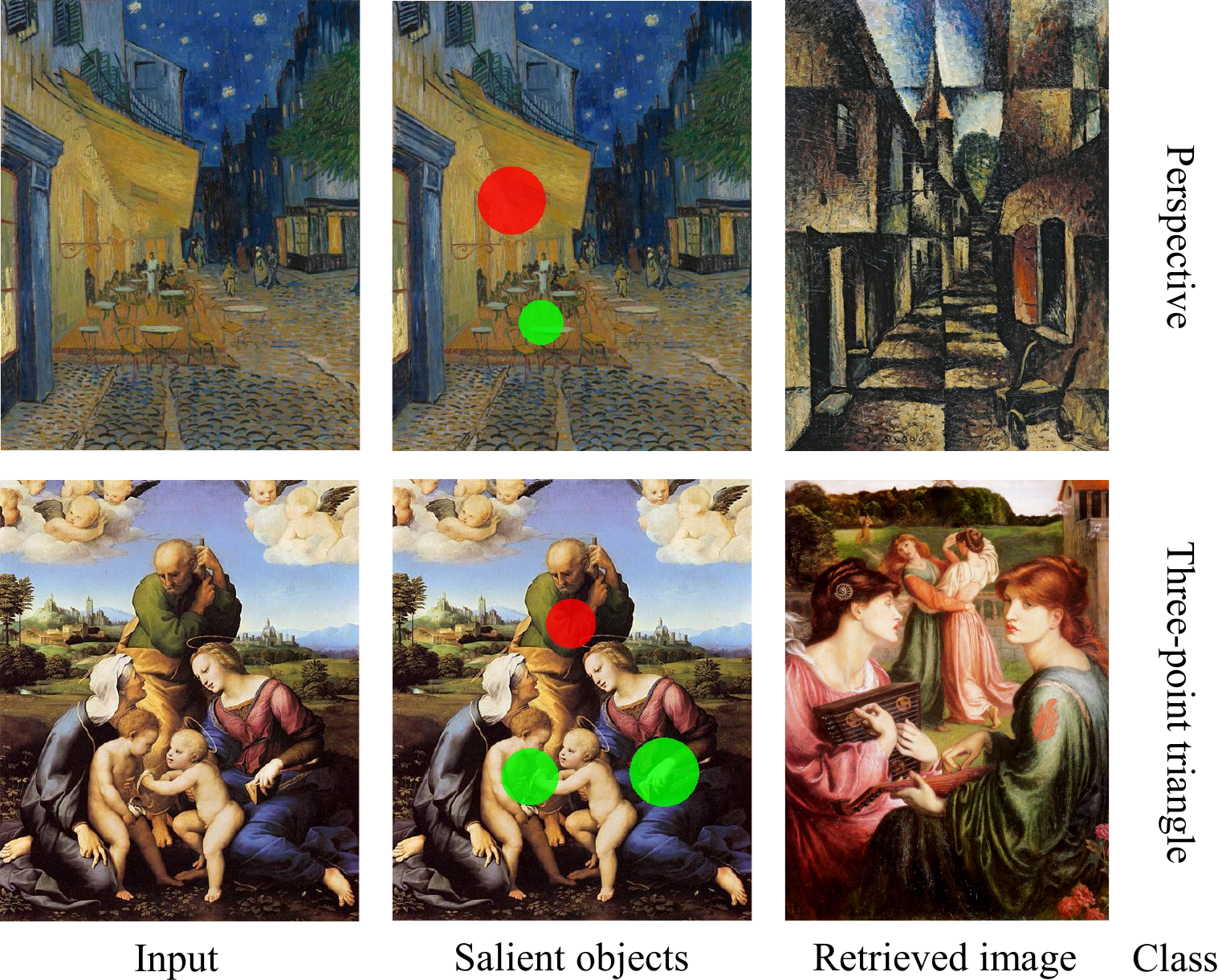}
    \caption{A well-structured compositional representation should generalize across diverse downstream tasks.}
    \label{fig:main_figure}
\end{figure}

\vspace{-11mm}
\begin{abstract}
  Composition, the deliberate arrangement of visual elements, is central to how meaning, emotion, and aesthetic quality are conveyed in artwork, yet it remains among the least formalized dimensions of visual understanding. Prior work highlights a persistent gap in learning meaningful compositional representations, attributing it to semantic bias and suggesting that human-inspired approaches may be key. 
  We compare two parallel paradigms for composition analysis: a human-inspired method grounded in perceptual grouping, and fine-tuned foundation models enabled by recent large-scale compositional datasets. The human-inspired approach uses object-centric models for region-level decomposition and a graph attention network to capture spatial relationships between elements. Both paradigms are evaluated on composition score/category prediction, compositional image retrieval, and visual saliency detection. With frozen encoders, the human-inspired method achieves competitive performance while remaining interpretable. When sufficient data enables fine-tuning, large self-supervised models outperform significantly, but at the cost of interpretability and cross-domain generalization. Code and pre-trained models are available on \href{https://github.com/FBehrad/Representation_learning_for_visual_composition}{\underline{GitHub}}.
\end{abstract}

\vspace{-8mm}
\section{Introduction}
Composition plays a fundamental role in conveying meaning, emotion, and aesthetic value in artwork by capturing the purposeful arrangement of visual elements that influences how viewers perceive an image \cite{arnheim1954art}. Developing computational models of composition could benefit applications such as aesthetic assessment, image retrieval, generative art, and cultural heritage, yet it remains among the most understudied aspects of visual understanding.

At the core of human perception of composition and spatial layout lie the mechanisms of perceptual grouping. Instead of analyzing an image as isolated pixels, the visual system organizes it into coherent regions \cite{wertheimer2012investigations, wagemans2015oxford} and derives structure from the spatial arrangement of these regions \cite{palmer1999vision}. 

Many computational approaches attempt to capture this process using graph attention networks (GAT) \cite{velivckovic2018graph} and transformers \cite{vaswani2017attention} to model relationships between image elements \cite{ghosal2022image, li2020composing, she2021hierarchical, zhao2020representation}. However, these methods typically operate on abstract feature representations produced by deep neural networks, rather than on semantically meaningful regions. As a result, the learned relationships are difficult to interpret and do not directly correspond to the region-based structure observed in human perception. In addition, these approaches struggle to separate compositional structure from semantic content \cite{zhao2025can}. 
Efforts to learn explicit region-based representations have been limited to bounding box regions \cite{zhao2025self}, which carry several limitations. First, the object detectors underlying this method are trained exclusively on natural images, making them difficult to transfer to artwork where annotated detection data is scarce. Second, operating on bounding boxes requires substantial feature engineering.

In this work, we adopt Object-Centric Learning (OCL), which decomposes images into structured region-level representations inspired by perceptual grouping principles \cite{locatello2020object}. Unlike bounding box methods, OCL models generalize across domains \cite{dittadi2022generalization} without requiring task-specific object detectors, making them naturally applicable to both photography and artwork. Building on these representations, we model inter-region relationships using a graph attention network, yielding a structured and interpretable compositional embedding.
We contrast this human-inspired paradigm against fine-tuned self-supervised foundation models, made feasible by recent advances in large-scale composition datasets \cite{jin2024apddv2, zhao2025can}. While such models do not follow perceptual grouping principles, they can achieve strong performance when sufficient labeled data is available. However, this comes at the cost of interpretability, domain generalizability, and dependence on large annotated data.

Beyond representation learning, we consider the broader utility of composition embeddings. Tasks such as compositional image retrieval and visual saliency prediction are often addressed separately using specialized models \cite{lin2024enhancing, le2020can, korkmaz2026visual}. In contrast, a representation that truly encodes composition can naturally support all of these tasks within a unified framework, without requiring additional supervision (\cref{fig:main_figure}). 

\section{Related work}
Composition in photography has been studied extensively, with works addressing both composition score prediction \cite{zhang2021image} and composition category classification \cite{lee2018photographic, zhao2025self}. The scale of available data has grown substantially with the introduction of the PICD dataset \cite{zhao2025can}, which enables large-scale neural network training for this task.
For artwork, while several aesthetic assessment datasets are available \cite{achlioptas2021artemis, yi2023towards, maerten2025lapis, amirshahi2014jenaesthetics}, those providing composition-specific labels remain scarce. The few that do focus on composition are too small (under 200 images) to support effective neural network training \cite{li2009aesthetic}. The most significant effort in this space is the APPDv2 dataset \cite{jin2024apddv2}, which comprises 10k painting images with attribute-level scores alongside overall aesthetic scores. Among these attributes is a composition and layout score, making it the largest available resource for composition analysis in artwork.

Early studies on composition analysis relied heavily on handcrafted features derived from photographic rules and were therefore limited to a narrow subset of predefined composition principles \cite{zhang2020inkthetics, dhar2011high}. 
Subsequent work shifted toward learning composition-aware representations directly using neural networks, leveraging supervision signals such as aesthetic or composition-related labels \cite{ozgun2023computational, lee2018photographic}. 
Complementary approaches incorporated saliency maps alongside predefined composition patterns to enable composition score distribution prediction \cite{zhang2021image}. 
Despite these advances, existing methods often exhibit limited cross-domain generalization (e.g., to artistic imagery) and rely on only weakly interpretable representations of composition.


In contrast to the aforementioned approaches, human visual perception operates in a different manner. 
Rather than processing a scene as a collection of raw pixels, the visual system first segments it into meaningful regions, then interprets the scene by reasoning about the spatial relationships between those regions.
Motivated by this principle, several models \cite{ghosal2022image, li2020composing, zhao2020representation, she2021hierarchical} have adopted graph attention networks or transformers to learn inter-region relationships.
In all cases, a neural network is first applied to extract high-level feature representations, which are then treated as graph nodes, and the model learns relationships between these abstract high-level features. While such approaches yield performance improvements, the nodes in the resulting graphs do not represent semantically meaningful regions. Therefore, these models do not faithfully reflect the region-level relational reasoning that characterizes human visual perception. Moreover, both graph-based models and transformer architectures exhibit computational complexity that scales with the size of the input representation. As a result, practical implementations often rely on strategies such as offloading intermediate graphs to disk to mitigate computational overhead \cite{ghosal2022image}.

The implementation closest to human composition understanding is proposed by Zhao et al. \cite{zhao2025self}, where regions are identified using bounding box attributes, such as size and position, obtained from an object detection algorithm. These region-level features serve as node features in the graph. Edges encode relational information between objects, capturing their relative size, distance, and direction. They also extract features using a neural network and employ a clustering model to identify structural primitives within the image.
While effective, the generalizability of this model to artwork remains questionable. The method is heavily dependent on bounding box information produced by an object detector trained on natural images, and directly applying such a model to a different domain carries a significant risk of performance degradation.

In a thorough analysis, Zhao \etal \cite{zhao2025can} identified that the composition embeddings created by current models are biased with substantial semantic content. 
Also, existing approaches demand large annotated datasets.
As mentioned earlier, humans distinguish compositions by explicitly recognizing compositional elements and their spatial arrangements. This suggests that architectures that explicitly model compositional elements and their topology can improve interpretability and data efficiency, while reducing semantic bias.

We can achieve this goal using object-centric representations, which have shown effectiveness in relational reasoning in prior work \cite{mosbach2024sold, ugadiarov2025relational}. 
By learning the relationships between these representations, we can explicitly capture compositional elements and their topology in a human-aligned manner. 
Moreover, operating on a small set of region-level representations (one per region) makes our model significantly more computationally efficient than prior approaches. Given that OCL models generalize well across domains \cite{dittadi2022generalization}, we aim to leverage this property to develop a single unified model capable of handling both photographs and artwork.
To the best of our knowledge, OCL models have not previously been explored in the context of composition understanding in photographs and artworks.
We compare this human-inspired paradigm with fine-tuned self-supervised foundation models, enabled by recent advances in large-scale composition datasets \cite{jin2024apddv2, zhao2025can}. While these models do not follow perceptual grouping principles, they can perform remarkably well given sufficient labeled data. This comparison addresses a key question:
\textit{Do we need to process visual information in a human-like manner to learn composition, or is a well-defined notion of composition combined with sufficient training data enough to unlock the potential of existing large-scale models?}

We further evaluate the quality of our compositional representations through their performance on downstream tasks.
Tasks such as composition-aware image retrieval and visual saliency prediction, the identification of the most salient element in a photograph or artwork, have traditionally been addressed in isolation, often requiring task-specific architectures \cite{lin2024enhancing} or specialized data such as eye-tracking annotations \cite{le2020can, korkmaz2026visual}. A high-quality composition embedding, however, naturally encodes the information needed for both tasks simultaneously, eliminating the need for dedicated models or costly data collection.
\vspace{-4mm}
\section{Compositional Representation Learning}
\label{sec:method}
\vspace{-2mm}
\begin{figure}
    \centering
    \includegraphics[width=0.85\linewidth]{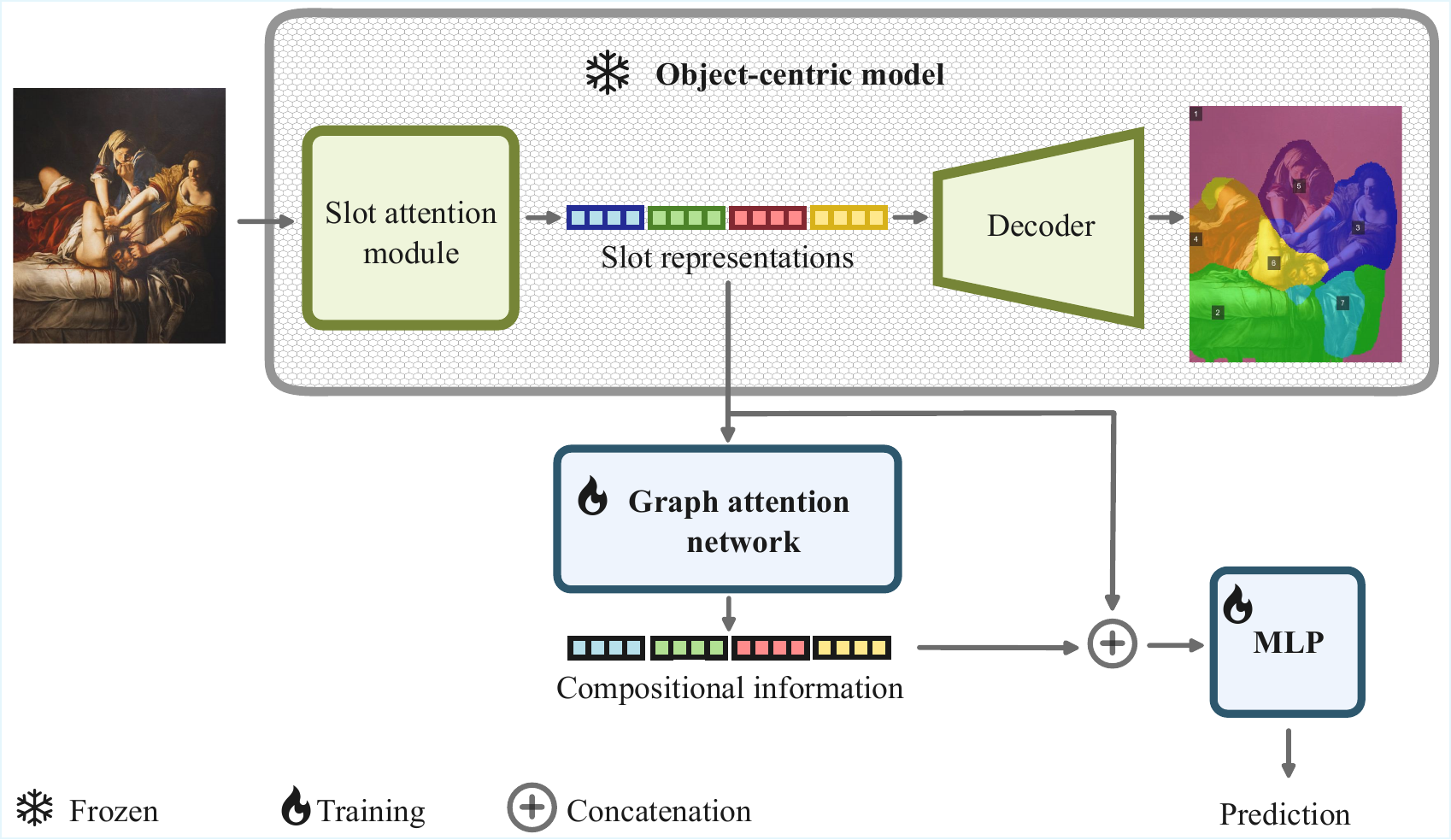}
    \caption{Overview of our architecture. Slot representations are extracted from the OCL model's slot attention module, where each slot corresponds to a meaningful image region. Then, GAT learns the relationships between these regions. The final feature representation is obtained by concatenating the slot features with the GAT output.}
    \label{fig:architecture}
\end{figure}
In this work, we compare compositional representations learned under two \\ paradigms: (1) fine-tuning foundation models and (2) explicitly modeling the relationship between object-centric representations (Fig. \ref{fig:architecture}).
Fine-tuning a foundation model trained in a self-supervised manner is a widely used approach in various topics. These models can achieve strong performance, especially when sufficient labeled data is available. However, they typically operate as black boxes and do not explicitly follow principles of human perceptual organization. 
In this work, we use object-centric learning to introduce an explicit inductive bias by decomposing an image into meaningful regions and modeling composition through relationships between these regions. Our goal is to understand which approach better captures compositional representations under different conditions.
\vspace{-4mm}
\subsection{Object-Centric Representation Learning}
\begin{figure}
    \centering
    \includegraphics[width=0.7\linewidth]{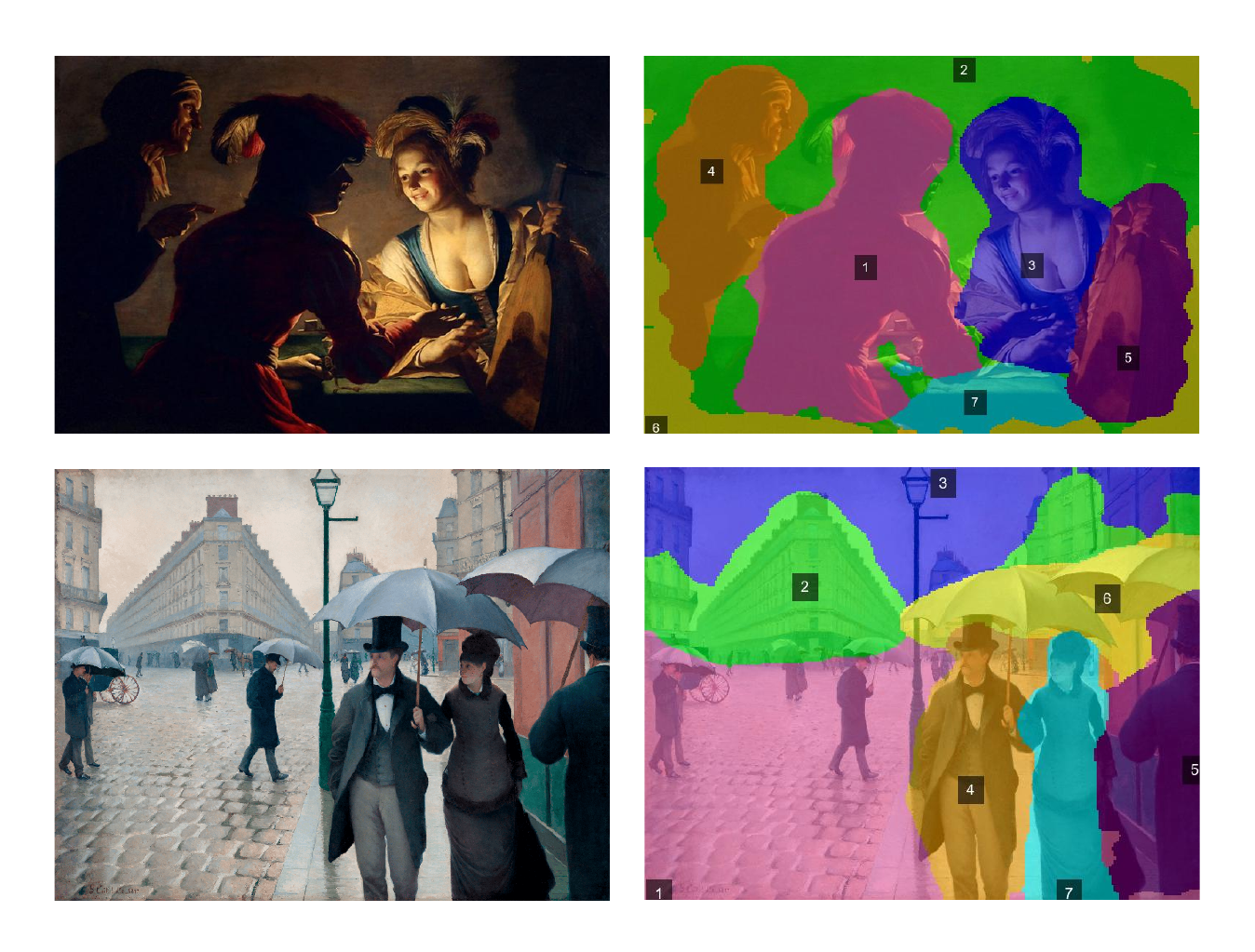}
    \caption{Segmentation output produced by our OCL baseline \cite{didolkar2025on} on artwork images, illustrating its ability to generalize beyond natural images and produce semantically meaningful regions across domains.}
    \label{fig:ocl_example}
\end{figure}

The key idea of object-centric learning is to represent an image as a set of meaningful regions, each with its \textit{own representation}.
We use a class of object-centric models called slot attention \cite{locatello2020object}, which takes image features from a pretrained backbone and clusters them into a fixed number of groups, called slots, using an attention-based mechanism. Each slot corresponds to a region in the image and produces a single representation. This process is conceptually similar to K-means clustering, while remaining fully differentiable. Importantly, this approach is close to the principle of perceptual grouping in human vision.

As shown in \cref{fig:ocl_example}, despite being trained on natural images, the model discovers semantically meaningful regions when applied to artwork, suggesting strong cross-domain generalization. We attribute this to the unsupervised nature of slot attention, which avoids overfitting to domain-specific appearance statistics. This makes OCL a suitable foundation for studying compositional representations across both photographs and artwork.

\subsection{Modeling relationships between regions}
To learn relationships between regions, we adopt graph attention networks \cite{velivckovic2018graph}, where each slot is considered a node, and edges represent relationships between slots. A key advantage of GAT is that it assigns different importance weights to different connections. This is important for compositional understanding, as not all regions contribute equally to the composition of an image. In addition, identifying salient elements requires emphasizing certain regions over others.
\vspace{-2mm}
\section{Experiments}
\subsection{Datasets}
The two primary datasets used in this study are PICD \cite{zhao2025can} and APDDv2 \cite{jin2024apddv2}, which are currently the largest available datasets for composition analysis in photography and artwork, respectively. 

The PICD dataset comprises 49k images distributed over 24 composition categories. In the original paper, the entire dataset was used for model evaluation; therefore, no official train/test split is provided. We applied stratified sampling to partition the data into train and test splits, reserving 20\% of each category for testing and the remainder for training. Further details are provided in Section 1 of the supplementary material.
The APDDv2 dataset comprises 10k painting images annotated with both overall aesthetic scores and attribute-level scores. One of these attributes is composition and layout, rated on a scale of 0 to 10, which we use as the supervision signal for our network.

Although OCL models generalize well to artwork in a zero-shot setting, we further fine-tune them on an artwork dataset (BAID dataset \cite{yi2023towards}) to investigate potential performance gains. Since OCL models are unsupervised, fine-tuning does not require segmentation labels. To evaluate the quality of the resulting segmentations, we use DRAM dataset \cite{cohen2022semantic}, the only existing artwork segmentation dataset. Full details of the fine-tuning procedure are provided in Section 2 of the supplementary material.
\vspace{-2mm}
\subsection{Implementation Details}
For all experiments, we train for 20 epochs using the AdamW optimizer with cosine annealing and a linear warmup of 2 epochs. The minimum learning rate is set to one-tenth of the maximum learning rate, with a single cosine cycle over the full training duration. When all feature encoders are frozen, we use the maximum learning rate of $1\times10^{-3}$. When fine-tuning Dinov2, we use the maximum learning rate of $1\times10^{-5}$. For multi-label classification on the PICD dataset, we use Binary Cross-Entropy with Logits loss (BCEWithLogitsLoss). For composition score prediction on the APDDv2 dataset, we use Mean Squared Error (MSE) loss.
We use FT-dinosaur \cite{didolkar2025on} with seven slots and a fixed slot initialization as our baseline OCL model. OCL is kept frozen throughout all experiments, as its performance is sufficient to justify foregoing fine-tuning. All experiments are conducted with a batch size of 32 on a single NVIDIA RTX A4500 GPU. 
\vspace{-2mm}
\subsection{Composition Category and Score Prediction}
\label{sec:sec4.3}
We evaluate the performance using (1) object-centric features only (\cref{fig:architecture}), and (2) Dinov2 features only.
Our settings allow us to directly compare structured object-centric representations with standard fine-tuning approaches.
\subsubsection{PICD dataset}
\begin{table}[tb]
  \caption{Performance comparison on \textbf{PICD test set}. {\color{RoyalBlue}Blue} and {\color{BrickRed}Red} show frozen and fine-tuned modules, respectively.
  }
  \label{tab:picd_results}
  \centering
  \begin{tabular}{@{}l|l|l|l|l@{}}
    \toprule
    Feature extractor & Precision (\%) & Recall (\%)& F1-score (\%)& Accuracy (\%)\\
    \midrule
    {\color{RoyalBlue}OCL} + {\color{BrickRed}GAT} &84.14&76.89&80.35&98.41\\
    {\color{RoyalBlue}OCL} + {\color{BrickRed}GAT} + {\color{RoyalBlue}Dinov2-b} & \underline{86.84}&78.03&82.20 & 98.57\\
    {\color{RoyalBlue}Dinov2-b}& 81.56 & 64.84& 72.25 & 97.90\\
    {\color{RoyalBlue}Dinov2-s}& 82.41 & 61.08 &70.16 & 97.81 \\
    \midrule
    {\color{BrickRed}Dinov2-b} & \textbf{88.66} & \textbf{88.21} & \textbf{88.44 }&\textbf{99.03}\\
    {\color{BrickRed}Dinov2-s} & 86.34 &\underline{86.31}&\underline{86.33}&\underline{98.85}\\
  \bottomrule
  \end{tabular}
\end{table}

As shown in \cref{tab:picd_results}, when all feature extractors are frozen, explicitly modeling composition through region-level relationships (OCL + GAT) yields considerable performance gains over using frozen Dinov2 features alone, across all metrics. However, composition is not always reducible to region relationships; factors such as color and contrast also play a defining role. This explains the additional performance improvement observed when frozen Dinov2 is used alongside OCL + GAT.

Fine-tuning Dinov2 consistently outperforms the frozen variants, suggesting that, given sufficient training data and a clear task design, a black-box network can surpass more structured, human-inspired approaches. 
The primary limitation of the fine-tuned Dinov2, however, is interpretability. While attention head visualization can highlight salient regions, it does not produce the clear, structured graph representation discussed in previous sections.

The performance of OCL + GAT is particularly noteworthy given that it achieves results comparable to fine-tuned Dinov2 variants while requiring only 990K trainable parameters, compared to approximately 22M and 86M parameters in Dinov2-s and Dinov2-b, respectively.

\vspace{-4mm}
\subsubsection{APDDv2 dataset}
\begin{table}[tb]
  \caption{Performance comparison on \textbf{APDDv2 test set}. {\color{RoyalBlue}Blue} and {\color{BrickRed}Red} show frozen and fine-tuned modules, respectively. OCL-BAID indicates OCL fine-tuned on the BAID dataset.
  }
  \label{tab:apdd_results}
  \centering
  \begin{tabular}{@{}l|l|l|l@{}}
    \toprule
    Model & PLCC (\%) & SRCC (\%)& Accuracy(\%) \\
    \midrule
    {\color{RoyalBlue}OCL} + {\color{BrickRed}GAT} & 54.45 & 53.27 & 81.65 \\
    {\color{RoyalBlue}OCL-BAID} + {\color{BrickRed}GAT} & 59.21 & 57.71 & 81.75\\
    {\color{RoyalBlue}OCL} + {\color{BrickRed}GAT} + {\color{RoyalBlue}Dinov2-b} & 67.84& 67.18 &\underline{83.57}\\
    {\color{RoyalBlue}OCL-BAID} + {\color{BrickRed}GAT} + {\color{RoyalBlue}Dinov2-b} & \underline{68.26}& \underline{67.19}& 82.56\\
    {\color{RoyalBlue}Dinov2-b} & 63.78 & 63.49& 82.66 \\
    {\color{RoyalBlue}Dinov2-s} & 56.25 &55.40 & 82.36\\
    \midrule
    {\color{BrickRed}Dinov2-b} & \textbf{68.94}&\textbf{67.48} &82.96\\
    {\color{BrickRed}Dinov2-s} & 64.50 & 62.69 & \textbf{85.08} \\
  \bottomrule
  \end{tabular}
\end{table}
\cref{tab:apdd_results} shows that fine-tuning OCL on BAID generally improves composition score prediction. However, unlike in \cref{tab:picd_results}, OCL + GAT does not outperform frozen Dinov2. Fine-tuning OCL on BAID allows it to surpass Dinov2-s but not Dinov2-b. Fine-tuning Dinov2 yields considerable gains in both PLCC and SRCC.

Analysis of the learned graph on APDDv2 reveals that the GAT fails to capture meaningful region relationships, producing a near-uniform attention distribution across edges (see supplementary, Section 3). This could be due to the fact that APDDv2 provides composition scores rather than composition categories. Composition scores are inherently more global, whereas predicting discrete composition categories may encourage the network to more actively attend to objects and their spatial relationships.

\textbf{Conclusion.} Our experiments on both datasets demonstrate that explicitly modeling the relationships between meaningful image regions is sufficient to achieve competitive performance in composition category classification. However, predicting a global composition score requires richer representations that capture broader image properties beyond inter-region relationships.


\vspace{-4mm}
\subsection{Downstream tasks}
Following Zhao \etal \cite{zhao2025can}, we evaluate our models on three downstream tasks designed to assess whether the learned representations capture genuine compositional structure or are biased toward semantic content. Since Zhao \etal \cite{zhao2025can} did not provide labels for these tasks, we construct dedicated test sets derived from our PICD test split.
\vspace{-4mm}
\subsubsection{Composition Feature Distinction (CFD)}
For this task, we construct 1,000 triplets, each consisting of an anchor image, a positive sample sharing the same composition category as the anchor, and a negative sample drawn from a different composition category. The underlying hypothesis is that images belonging to the same composition category should yield closer representations than images from different categories.
We measure performance using triplet accuracy, defined as the proportion of triplets for which the distance between the anchor and the positive sample is smaller than the distance between the anchor and the negative sample:
\begin{equation}
\text{Accuracy} = \frac{1}{N} \sum_{i=1}^{N} \mathbf{1} \left[ d(a_i, p_i) < d(a_i, n_i) \right]
\label{equ:eq1}
\end{equation}
where $N$ is the total number of triplets, $a_i$, $p_i$, and $n_i$ denote the anchor, positive, and negative embeddings respectively, and $d(\cdot, \cdot)$ denotes the cosine distance between two representations.
\vspace{-4mm}
\subsubsection{Robustness to Semantic Interference (SI)}
The second task evaluates \\whether the learned representations remain sensitive to composition rather than being driven by semantic content. We first predict the class of each image using ViT-B \cite{dosovitskiy2021an}, as the PICD dataset does not provide semantic labels. We then construct 1,000 triplets, where each anchor is paired with a positive sample sharing the same composition category but depicting different semantic content, and a negative sample sharing the same semantic content as the anchor but belonging to a different composition category. Again, the hypothesis is that images with the same composition should yield closer representations than images with the same semantics but different composition, thereby confirming that the model has not conflated compositional and semantic information.
Performance is measured using \cref{equ:eq1}.
\vspace{-4mm}
\subsubsection{Compositional retrieval}
For this task, we use the full PICD test set of 9,600 images. For each anchor image, the goal is to retrieve images belonging to the same composition category from the remaining test images. We report performance using Precision@$K$ for $K \in \{1, 5, 10\}$ and mean average precision (mAP).
\vspace{-8mm}
\subsubsection{Performance on downstream tasks}
\begin{table}[tb]
  \caption{Performance comparison on the \textbf{CFD} and the \textbf{SI} task measured in accuracy (\%).  {\color{RoyalBlue}Blue} and {\color{BrickRed}Red} show frozen and fine-tuned modules, respectively. All models are trained on the PICD dataset.
  }
  \label{tab:CFD_SI}
  \centering
  \begin{tabular}{@{}l|l|l@{}}
    \toprule
    Model & CFD & SI \\
    \midrule
    {\color{RoyalBlue}OCL} + {\color{BrickRed}GAT} & 86.9 & 70.0 \\
   {\color{RoyalBlue}OCL} + {\color{BrickRed}GAT} + {\color{RoyalBlue}Dinov2-b}  & 70.8 & 28.6\\
     {\color{RoyalBlue}Dinov2-b}  & 68.8 & 28.2 \\
     {\color{RoyalBlue}Dinov2-s} & 67.5 & 31.3 \\
     \midrule
     {\color{BrickRed}Dinov2-b} & \textbf{94.3} & \textbf{94.7} \\
     {\color{BrickRed}Dinov2-s} & \underline{93.7} & \underline{91.0} \\
     
  \bottomrule
  \end{tabular}
\end{table}

\begin{table}[tb]
  \caption{Performance comparison on the \textbf{composition retrieval} task.  {\color{RoyalBlue}Blue} and {\color{BrickRed}Red} show frozen and fine-tuned modules, respectively. All models are trained on the PICD dataset. All metrics are reported as a percentage (\%).
  }
  \label{tab:picd_retreival}
  \centering
  \begin{tabular}{@{}l|l|l|l|l@{}}
    \toprule
    Model & mAP& Precision@1& Precision@5& Precision@10\\
    \midrule
    {\color{RoyalBlue}OCL} + {\color{BrickRed}GAT}  & 32.08 & 69.64 & 65.06& 62.44\\
    {\color{RoyalBlue}OCL} + {\color{BrickRed}GAT} + {\color{RoyalBlue}Dinov2-b}& 19.13 & 55.49 & 49.56& 46.25 \\
     {\color{RoyalBlue}Dinov2-b}& 18.60 & 55.14 & 49.26 & 45.99\\
    {\color{RoyalBlue}Dinov2-s}& 20.19 & 57.36 & 51.67 & 48.46\\    
     \midrule
     {\color{BrickRed}Dinov2-b} &  \textbf{82.18}&\textbf{86.00} &\textbf{86.00} &\textbf{86.00} \\
     {\color{BrickRed}Dinov2-s} & \underline{79.20} & \underline{83.73} & \underline{83.77} & \underline{83.84}\\
     
  \bottomrule
  \end{tabular}
\end{table}

As shown in \cref{tab:CFD_SI} and \cref{tab:picd_retreival}, when all feature extractors are frozen, OCL + GAT substantially outperforms both Dinov2-s and Dinov2-b on all tasks. This is expected, as both Dinov2 variants are pretrained for image classification and therefore produce predominantly semantic representations, making them poorly suited for capturing compositional structure in a frozen state.
However, fine-tuning Dinov2 leads to considerable performance improvements on all tasks. This is a notable finding, as it appears to contradict the conclusions of prior work \cite{zhao2025can}, which generally characterized existing models as semantically biased and incapable of capturing compositional information. 

Our results suggest that, given \textit{sufficient data} and a \textit{well-defined composition analysis objective}, these models can adapt and learn representations that are genuinely task-relevant rather than purely semantic.
We hypothesize that the limited performance observed in prior work reflects the nature of the training objective rather than inherent limitations of the models themselves. Many of the models evaluated in prior work \cite{zhao2025can} were not trained specifically for composition classification, but rather for related yet distinct tasks such as aesthetic assessment \cite{ghosal2022image, yi2023towards, he2023eat, she2021hierarchical} or image cropping \cite{zheng2022grid, su2024spatial, li2020composing, Hong_2021_CVPR}. A clearly defined composition-related objective appears to be key to enabling these models to move beyond semantic bias.

Nevertheless, when interpretability is required or when feature extractors must remain frozen due to computational limitations, human-inspired architectures such as OCL + GAT offer a clear advantage. 

\begin{figure}[t]
    \centering
    \includegraphics[width=0.7\linewidth]{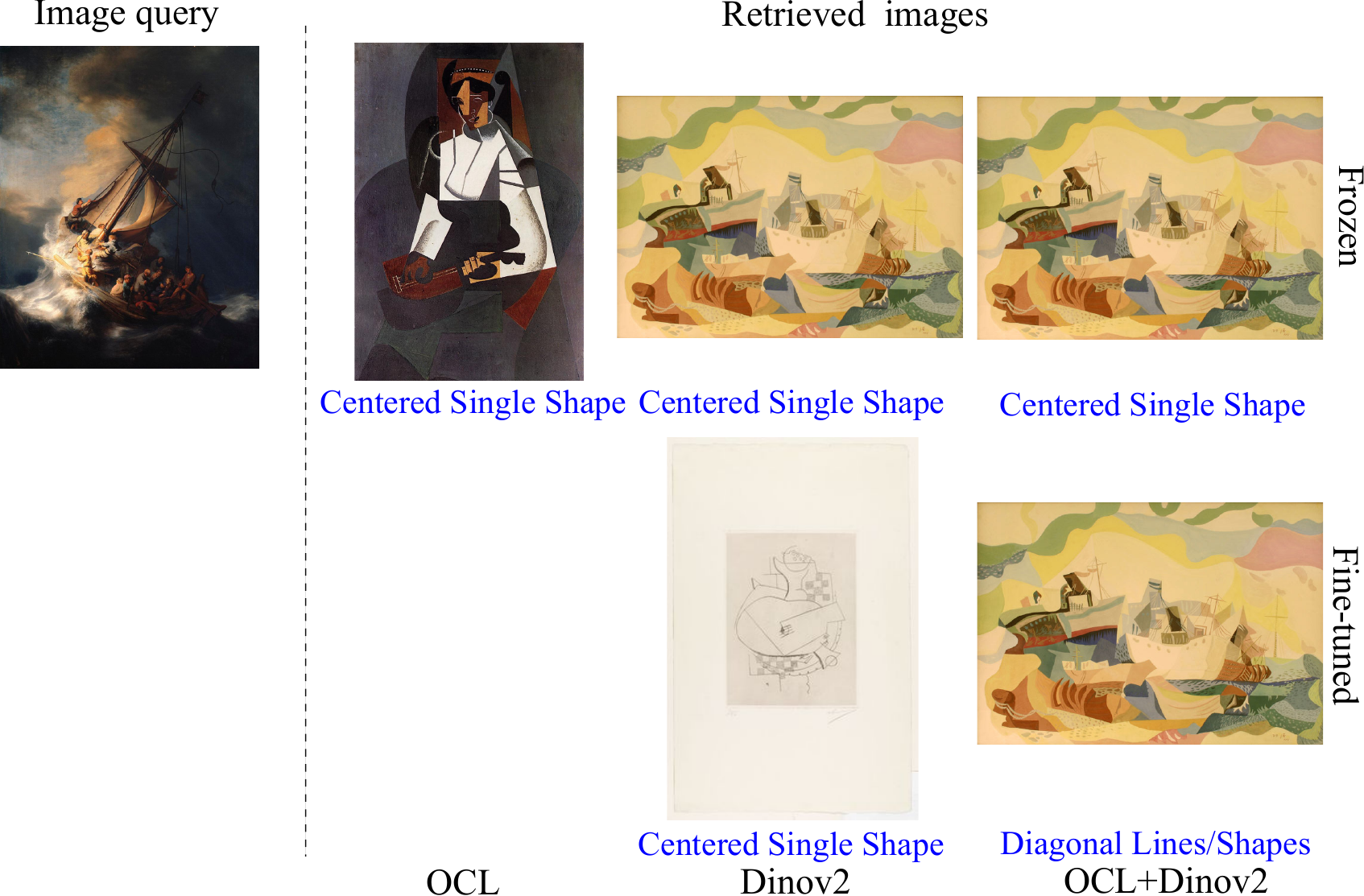}
    \caption{This figure shows the query image and the top retrieved images from each model, along with each model’s prediction. The results indicate that retrieval is strongly driven by the predicted class.}
    \vspace{-6mm}
    \label{fig:retrival}
\end{figure}

\cref{fig:retrival} shows visual examples of compositional image retrieval. We observe that image retrieval is strongly driven by the predicted composition class, with models consistently retrieving images that match this class (e.g., “centered single shape” retrieves similar compositions). OCL + GAT and fine-tuned Dinov2 are more aligned with composition and less semantically biased, while other models are more influenced by semantics. Additional examples are provided in Section 4 of the supplementary material.

When applying a model pretrained on PICD to artwork retrieval, quantitative evaluation is hindered by the absence of composition category labels in APDDv2 and the inherently less well-defined compositional categories in artwork. Qualitative results, however, suggest that the learned representations remain visually reasonable, though they expose clear limitations in cross-domain transfer. For instance, in \cref{fig:retrival}, although the query image and the retrieved image by OCL share the same composition category (centered single shape), their visual compositions differ significantly. 
This highlights that the \textit{definition of composition labels} is critical, as models learn these labels.
In addition, based on the visual examples, it is questionable if semantic information should be fully removed or partially retained. 
\vspace{-4mm}
\subsubsection{Visual saliency detection}
Although fine-tuned Dinov2 models considerably outperform OCL + GAT, they carry notable limitations. As a black-box model, Dinov2 is less interpretable. In contrast, our OCL + GAT framework provides a unified, interpretable architecture. In our architecture, OCL produces region-level representations, while the GAT explicitly models the relationships between these regions. Crucially, this design enables a capability that Dinov2 cannot provide, namely, the identification of the importance of different image regions with respect to composition.

\begin{figure}
    \centering
    \includegraphics[width=0.8\linewidth]{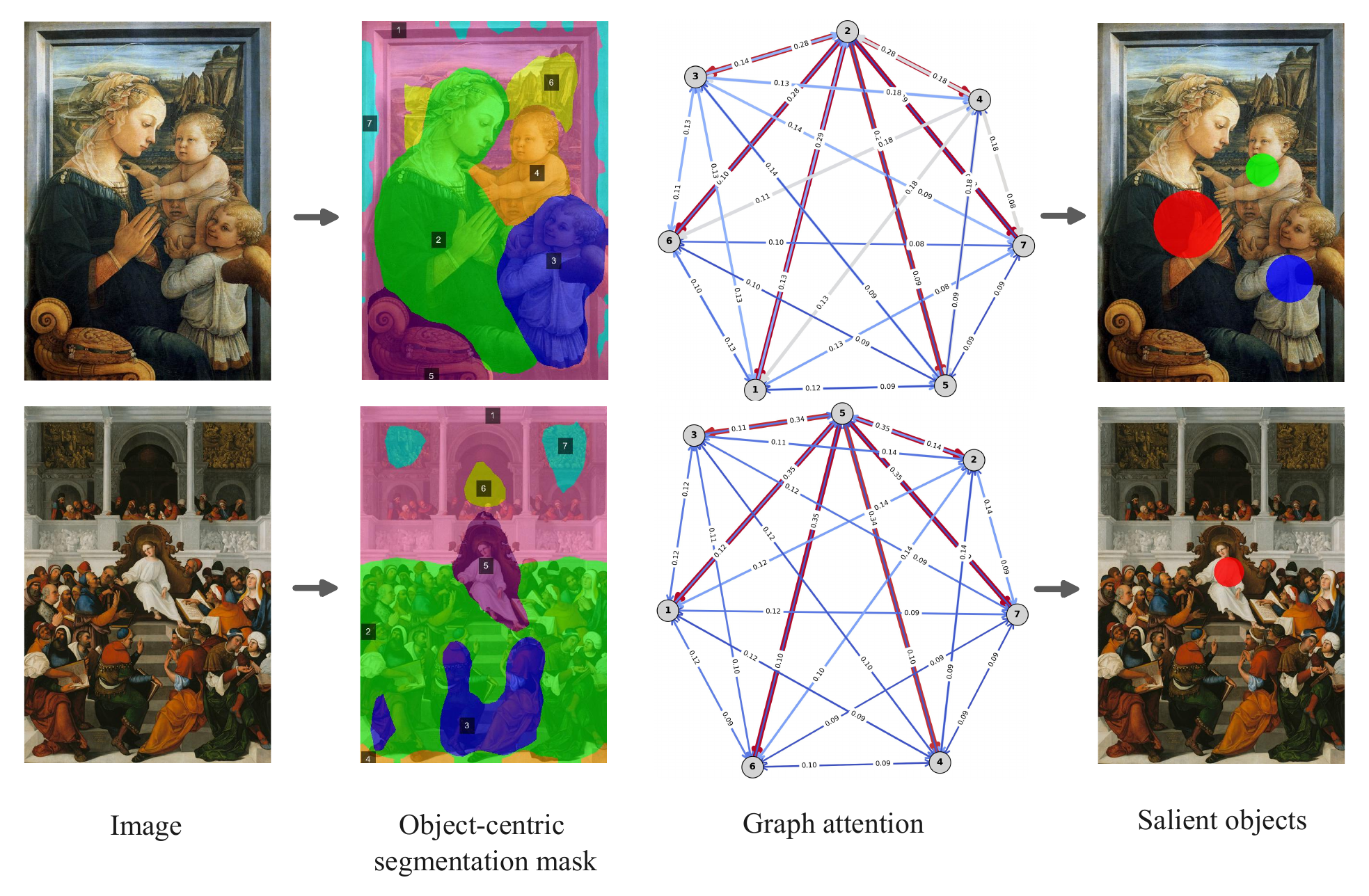}
    \caption{Visual saliency detection derived from the learned GAT. The importance of each region is computed as the sum of its outgoing edge weights. The most important regions are shown in red, green, and blue, respectively. If a large gap exists between consecutive importance scores, only the regions above the gap are retained. Node indices in the segmentation mask correspond to 
    graph nodes, and circle sizes reflect the size of the segmented region.}
    \vspace{-6mm}
    \label{fig:visual_saliency}
\end{figure}

Saliency is derived directly from the graph structure. For each node, we compute an importance score as the sum of its outgoing edge weights to identify the node with the greatest influence on other regions. The top-k nodes by importance score are then selected as the most salient regions. Alternatively, saliency can be determined dynamically by thresholding on the difference between consecutive node importance scores, retaining only those nodes above a predefined gap (0.2 in our study). This allows the number of salient regions to vary across images rather than being fixed. \cref{fig:visual_saliency} shows two illustrative examples. In the first image, the most salient region is the woman, highlighted in red, followed by a second region in green and a third in blue. In the second image, only a single salient region is identified, demonstrating the flexibility of the dynamic thresholding approach.

\begin{figure}
    \centering
    \includegraphics[width=0.7\linewidth]{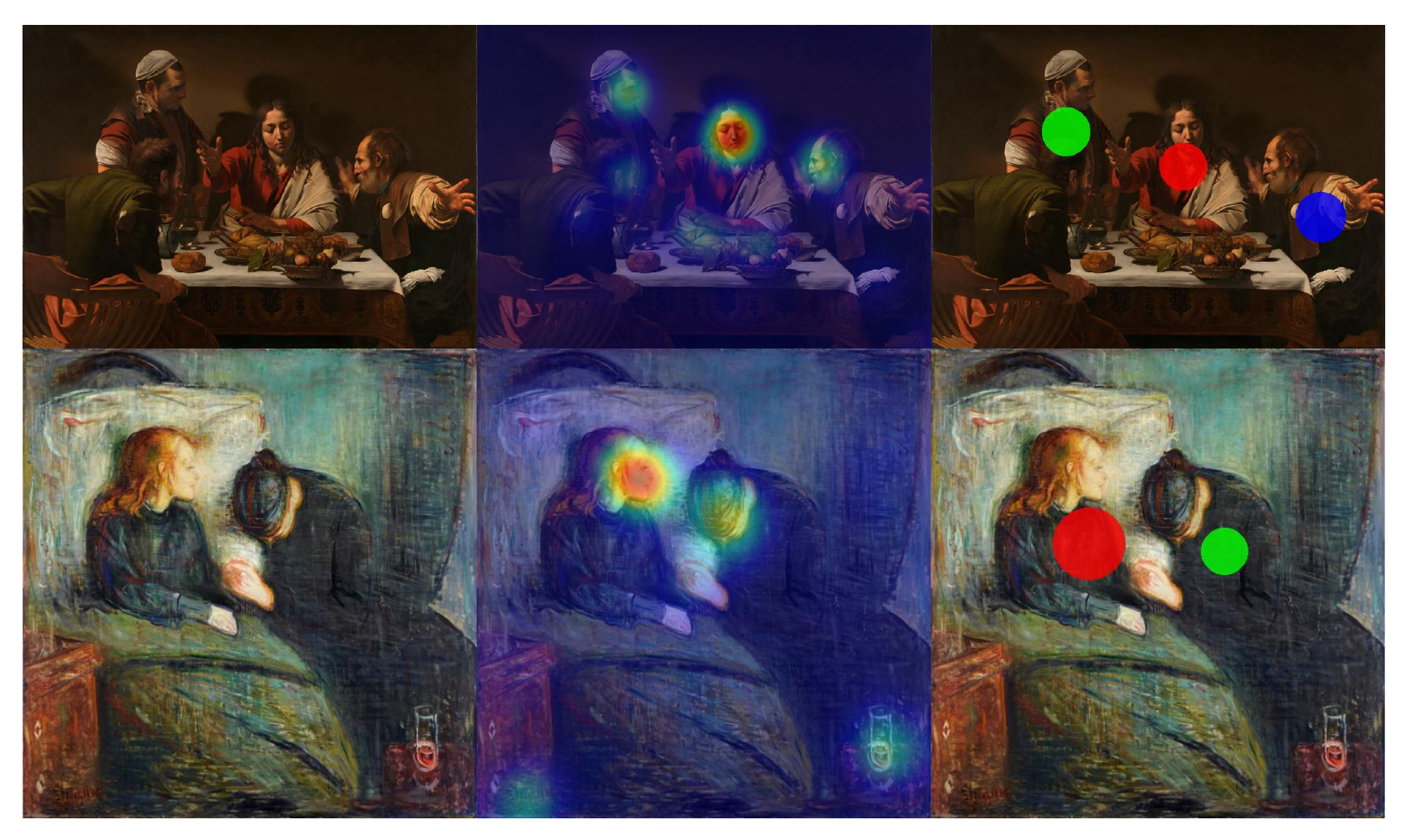}
    \caption{Comparison between visually salient regions detected by our model and human fixation patterns from eye-tracking data \cite{wagemans2026saccade}. Our model correctly identifies the most salient regions and their relative importance ordering.}
    \vspace{-7mm}
    \label{fig:eye_tracking}
\end{figure}

To assess our saliency estimation, we compare model outputs against eye-tracking data from \cite{wagemans2026saccade}. Direct quantitative comparison is not straightforward because our method identifies salient regions per image rather than a dense saliency map. In addition, OCL segments images at a coarse level (e.g., capturing whole objects such as a person rather than finer parts like the face), making precise alignment with detailed fixation patterns infeasible.
Despite this, as shown in \cref{fig:eye_tracking}, our model consistently identifies the regions that human observers fixate on most. Spatial discrepancies between predicted regions and fixation points are expected and largely attributable to the coarse granularity of OCL masks.

Moreover, the ranking of region importance produced by our model aligns with the order of visual attention observed in the eye-tracking data. This correspondence holds across the majority of examples, suggesting that our model reliably identifies salient objects in a manner consistent with human visual attention. See additional examples in Section 5.2 of the supplementary material.
\vspace{-6mm}


\vspace{-6mm}
\subsection{Attention Faithfulness}
\vspace{-6mm}
\begin{figure}
    \centering
    \includegraphics[width=0.5\linewidth]{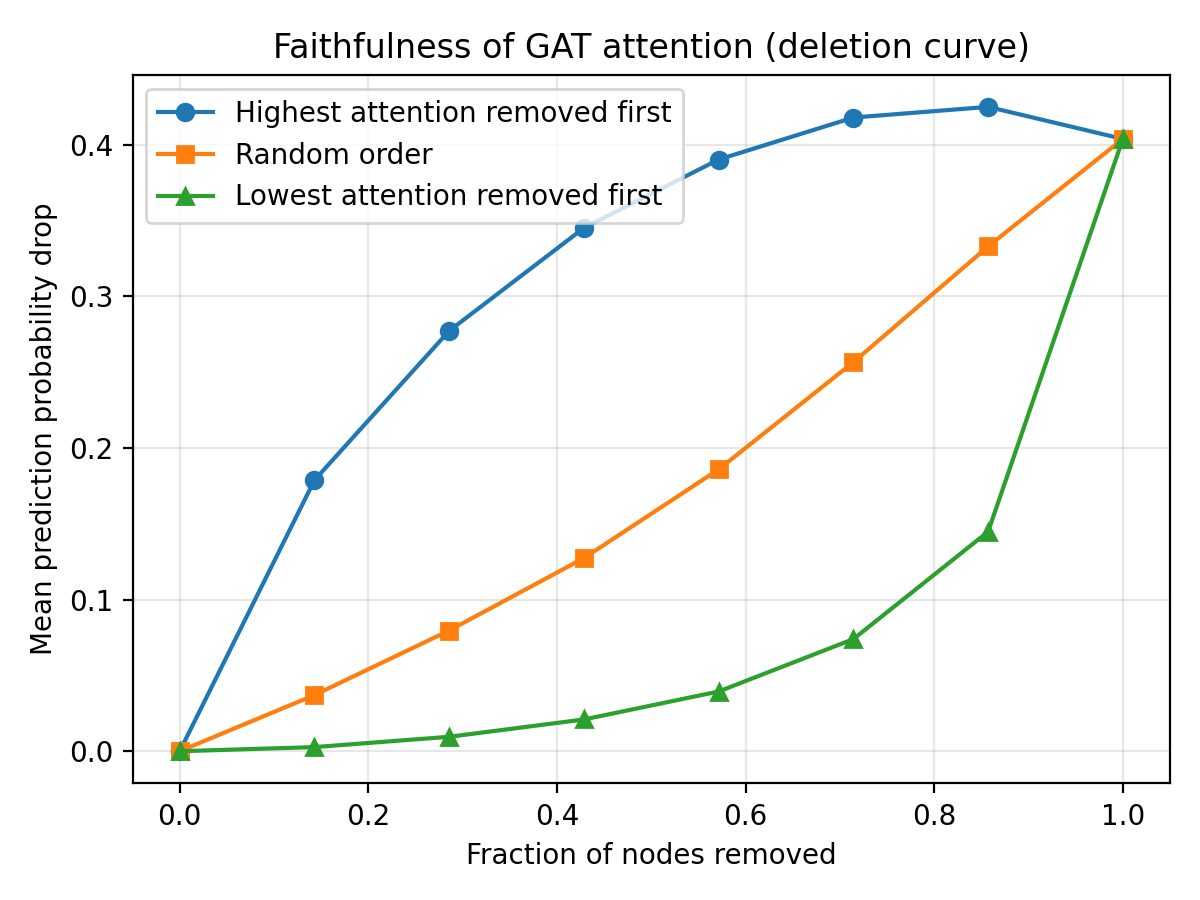}
    \caption{
    Faithfulness of GAT attention evaluated using a deletion-curve analysis on the full PICD test set. Removing slots with the highest outgoing attention weights leads to a faster decrease in prediction confidence than random removal, whereas removing low-attention slots first has little effect until most nodes are gone.}
    \vspace{-6mm}
    \label{fig:deletion_curve}
\end{figure}

Until now, we have assumed that GAT attention weights reflect the importance of image regions for predicting composition. We test this assumption using a deletion-curve analysis. This evaluation is adapted from the deletion metric introduced by Petsiuk et al.~\cite{petsiuk2018rise}, which measures explanation faithfulness by progressively removing features according to their importance and tracking the resulting change in the model's prediction. In our setting, the features are the slots represented by GAT nodes.
For each test image, we progressively remove GAT nodes in three different orders: from highest to lowest outgoing attention weight, from lowest to highest outgoing attention weight, and in a random order averaged over three permutations. At each step, we measure the drop in the model's predicted class probabilities relative to the original, unablated prediction.

Fig.~\ref{fig:deletion_curve} shows the results for OCL + GAT on the full PICD test set. Removing the highest-attention nodes causes the prediction confidence to decrease substantially faster than random removal. In contrast, removing the lowest-attention nodes has a smaller effect on the prediction. The area between the high-attention and random curves is 0.1449, while the area between the random and low-attention curves is 0.1040. Thus, the model is more sensitive to the removal of regions that receive high attention from the GAT.
The effect is already visible after removing a single node. The high-attention condition produces a prediction-confidence drop approximately $5\times$ larger than random removal. Overall, the ordering of the curves, high $>$ random $>$ low, shows that the regions assigned higher attention are, on average, more important for the model's prediction than those assigned lower attention.

As an additional test, we evaluated the model using only the three most- or least-important slots. Keeping the top-3 slots results in an F1 score of 73.83\%, compared with 36.87\% when only the bottom-3 slots are retained. This result provides further evidence that the attention weights identify regions that are important for the model's prediction. More details are provided in Section 6 of the supplementary material.

\vspace{-3mm}
\section{Limitation and Future Work}
\begin{figure}[t]
    \centering
    \includegraphics[width=0.8\linewidth]{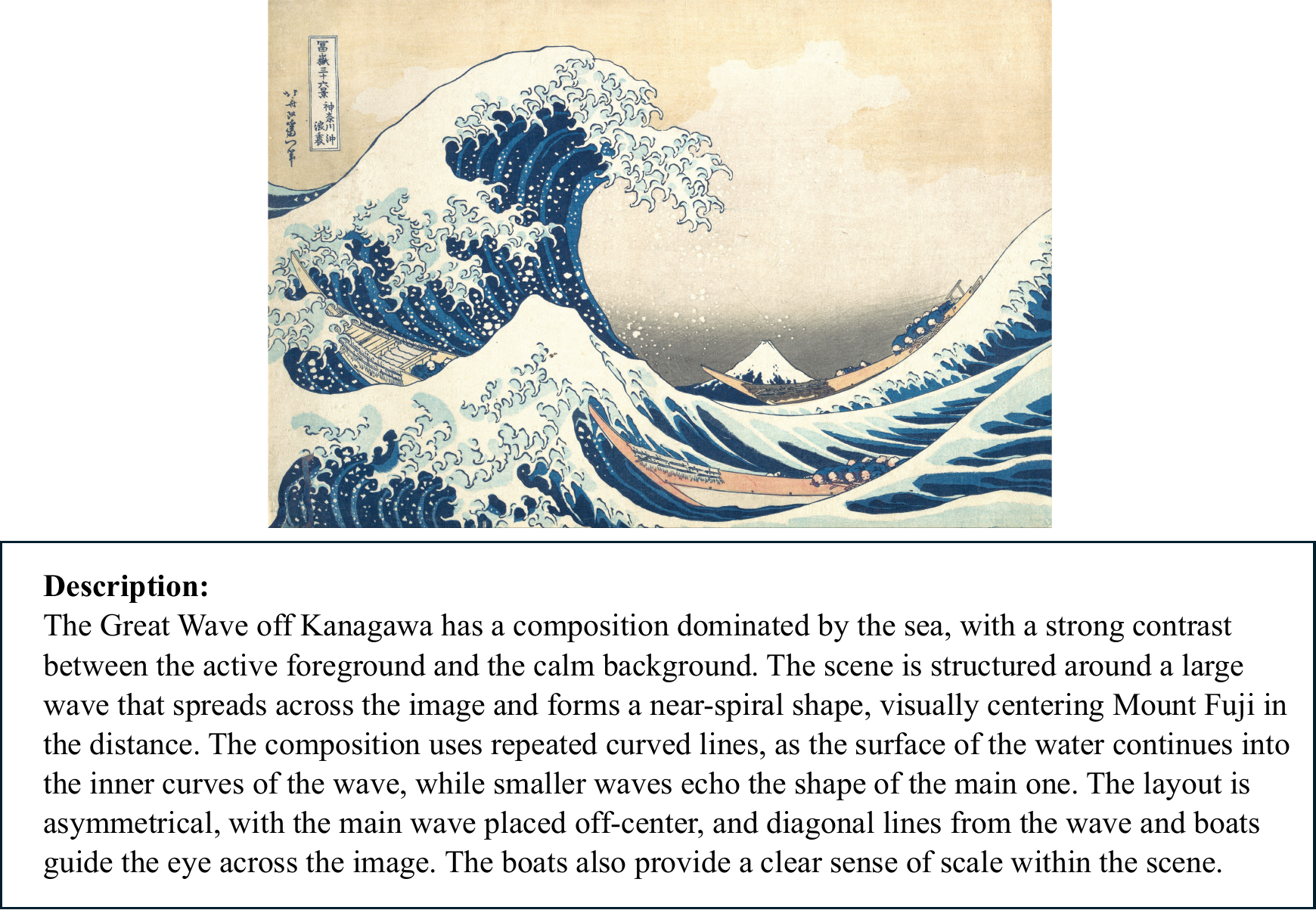}
    \caption{The Great Wave off Kanagawa and its corresponding composition description.}
    \vspace{-6mm}
    \label{fig:composition_description}
\end{figure}

\vspace{-1mm}
While numerous datasets exist for studying composition in photography, datasets for analyzing artwork remain limited. Most available artwork composition datasets provide only a single global score, which is insufficient to capture relationships between image regions (\cref{sec:sec4.3}) and fails to reflect the inherently multidimensional nature of artistic composition. As a result, developing rich and fine-grained annotations is essential for advancing compositional understanding. One promising direction is the use of compositional descriptions. Prior work has explored this idea in photography, where some studies focus directly on composition \cite{yuan2026towards, you2025photoframer}, while others incorporate it as part of broader aesthetic descriptions \cite{huang2024aesexpert, vera2022understanding, qi2025photographer}. In contrast, such efforts in the domain of artwork are still scarce. Existing approaches primarily describe overall content \cite{stefanini2019artpedia, bai2021explain, lu2022artcap} or general aesthetics \cite{jin2024apddv2}, without offering detailed compositional explanations. To date, no dataset systematically captures compositional structure in artworks through dedicated descriptions. 
\cref{fig:composition_description} illustrates an example of such a compositional description, highlighting how structured annotations can support richer compositional learning. Wikipedia provides detailed composition descriptions for many artworks. Given that current LLMs are trained on internet-scale data, they represent a promising source for generating such annotations automatically.

Although our model generalizes reasonably from photography to artwork, particularly for categories such as rule of thirds and perspective, photographic composition concepts are often too rigid to capture the more complex and dynamic structures found in artwork, limiting cross-domain transfer. Furthermore, compositionally relevant categories such as symmetry are absent from PICD, constraining the scope of what the model can learn.

Finally, our model lacks hierarchical attention, leading to coarse predictions in tasks such as saliency detection (e.g., identifying a person but not their most salient part, such as the face in \cref{fig:eye_tracking}). Future work could incorporate hierarchical relational modeling.

\vspace{-5mm}
\section{Conclusion}
We investigated the gap between human compositional understanding and existing computational models by comparing a human-inspired pipeline, combining object-centric region decomposition with graph attention networks, against fine-tuned large foundation models. Our results show that when sufficient annotated data is available, fine-tuned foundation models achieve superior task-specific performance. However, our human-inspired approach offers compelling advantages in interpretability, computational efficiency, and cross-domain generalization.
Notably, explicitly modeling meaningful regions and their relationships proves sufficient for composition category prediction, achieving performance comparable to large foundation models.  However, this approach yields lower performance on composition score prediction, as the overall composition score in an artwork is influenced by many factors beyond inter-region relationships.




\vspace{-5mm}
\section*{Acknowledgements}
Funded by the European Union (ERC AdG, GRAPPA,
101053925, awarded to Johan Wagemans) and the Research Foundation-Flanders (FWO, 1159925N, awarded to Fatemeh Behrad). 
We would like to thank Maarten Leemans and Doreen Hii for sharing the eye-tracking data from their study.

%
%
\bibliographystyle{splncs04}
\bibliography{main}

\title{Supplementary materials: Learning visual representations for compositional analysis of artworks and photographs}
\titlerunning{Learning visual representations for compositional analysis}

\author{Fatemeh Behrad\orcidlink{0000-0003-2629-0854} \and
Tinne Tuytelaars
\orcidlink{0000-0003-3307-9723} \and
Johan Wagemans\orcidlink{0000-0002-7970-1541}}


\institute{
KU Leuven University\\ Belgium
}

\maketitle

\section{PICD train/test split}
\begin{figure}
    \centering
    \includegraphics[width=0.85\linewidth]{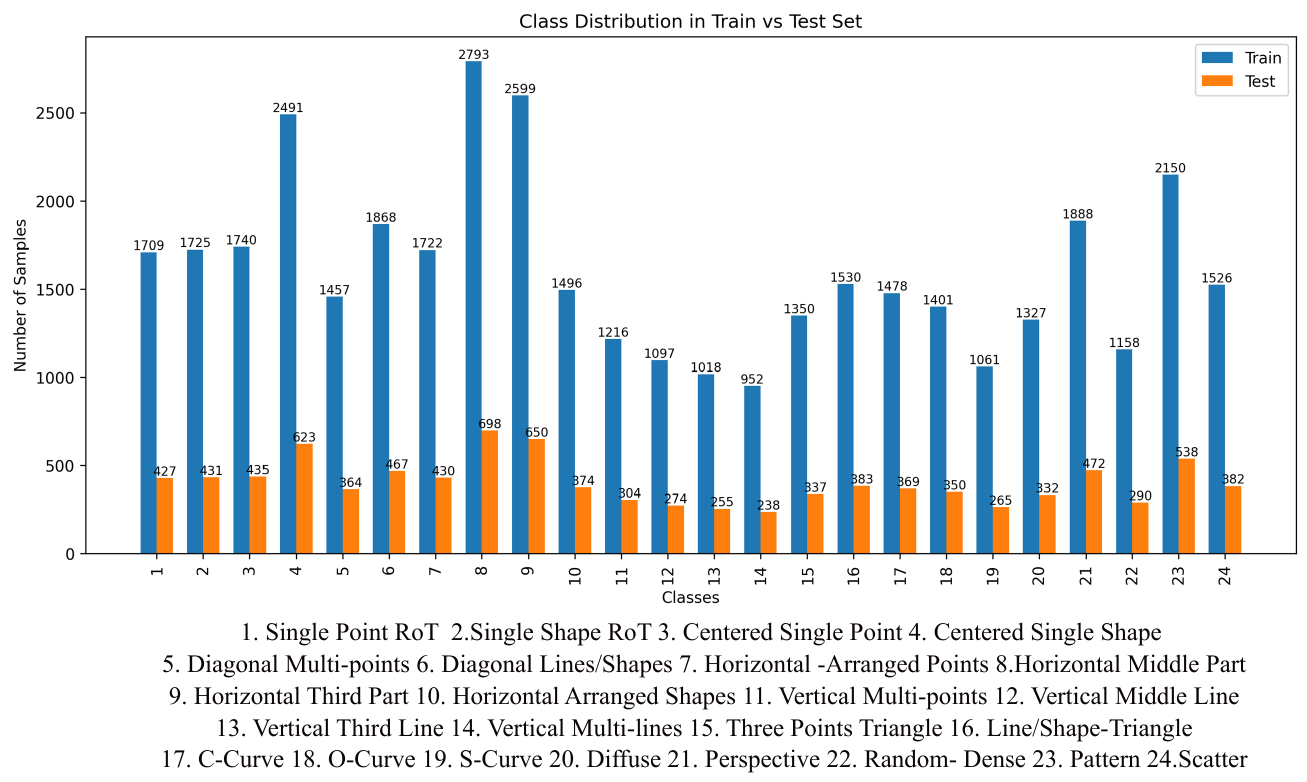}
    \caption{Class distribution of the PICD dataset across training and test splits.}
    \label{fig:picd}
\end{figure}

Zhao et al. \cite{zhao2025can} used the entire PICD dataset for evaluating existing models, providing no official train/test split. Given the scale of the dataset, we repurpose it for training by constructing our own split as follows.
Since some images belong to multiple composition categories, we first identify multi-label combinations that occur only once in the dataset. We assign them exclusively to the test set.
We then apply stratified sampling over the remaining images, partitioning the data such that each composition category is proportionally represented in both splits. Specifically, 20\% of the images in each category are reserved for testing, with the remainder used for training and validation. The resulting class distribution is shown in \cref{fig:picd}.

\vspace{-4mm}
\vspace{-5mm}
\section{Object-Centric Model Fine-Tuning}
Slot representations are initialized by randomly sampling from a learned Gaussian distribution. These slots are then passed through a competitive attention module, in which slots compete to capture different subsets of the input tokens, effectively performing a form of soft clustering over the input features. The resulting slots are passed to a decoder, which reconstructs the image or input features from these representations. The model is trained end-to-end using this encoder-decoder objective, without any supervisory labels \cite{locatello2020object, didolkar2025on}. This unsupervised formulation lends itself to the task of object discovery, where the goal is to decompose an image into its constituent regions without any labeled supervision. The quality of the discovered regions is evaluated using metrics such as FG-ARI, mBO, and mIoU, which measure the correspondence between the learned slot assignments and ground truth segmentation masks.

A known limitation of random slot initialization is inconsistency across runs. In other words, different random seeds yield different slot assignments, making it difficult to learn stable and reproducible representations.
However, it is necessary to have consistent slots in our task as we want to learn their relationship with GAT. To address this, following previous work \cite{manasyan2025temporally}, we replace random sampling with learned slot initializations. In this approach, rather than sampling slot initializations from a learned Gaussian distribution, slots are treated as directly learnable parameters, ensuring consistent and deterministic outputs across runs.

Building on this, we train FT-dinosaur using learned slot initializations \cite{manasyan2025temporally} using the same hyperparameters as \cite{didolkar2025on}. We then fine-tune it on the BAID dataset  \cite{yi2023towards}. The BAID dataset, which was originally collected for aesthetic assessment, comprises images collected from an online platform.
We evaluate the resulting segmentation quality on the DRAM dataset \cite{cohen2022semantic}, which is sourced from WikiArt\footnote{\url{https://www.wikiart.org/}}, ensuring no overlap between training and test data.

\begin{figure}
    \centering
    \includegraphics[width=0.85\linewidth]{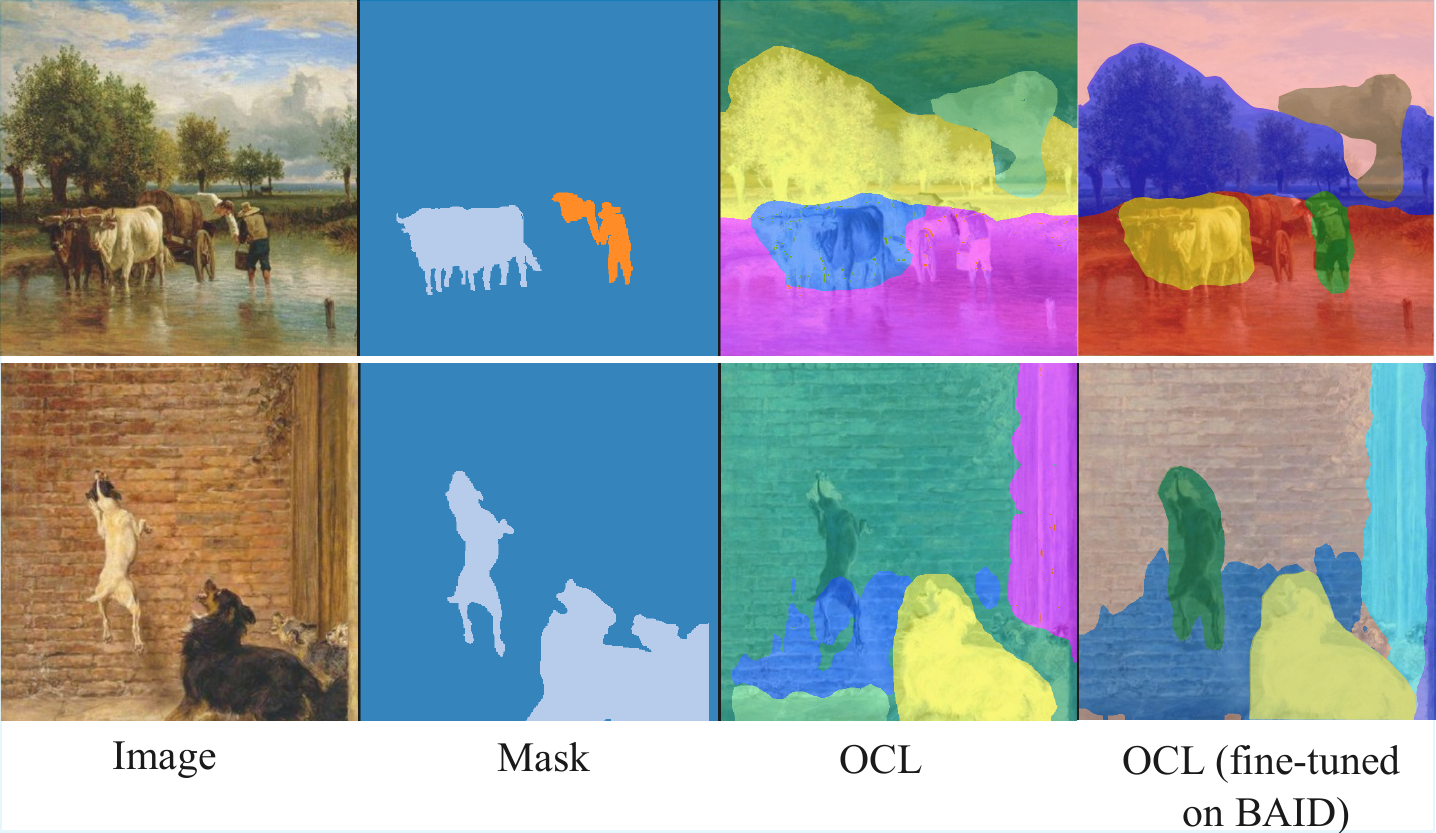}
    \caption{Fine-tuning our OCL baseline (FT-dinosaur + learnable slot initialization) on the BAID dataset improves the object discovery in artwork.}
    \label{fig:baid_finetning}
\end{figure}

As shown in \cref{tab:baid_fine_tuning}, fine-tuning yields a slight improvement on the DRAM dataset, and a considerably larger improvement on the downstream composition task (Section 4.3). As shown in \cref{fig:baid_finetning}, fine-tuning on the BAID dataset improves the object discovery in the paintings (e.g., the man in the field and the dog on the wall).

\begin{table}[tb]
  \caption{Performance on the DRAM dataset. All models have a learnable slot initialization. {\color{RoyalBlue}Blue} and {\color{BrickRed}Red} show frozen and fine-tuned model on BAID, respectively.
  }
  \label{tab:baid_fine_tuning}
  \centering
  \begin{tabular}{@{}l|l|l|l@{}}
    \toprule
    Model & mBO& FG-ARI& mIoU\\
    \midrule
    {\color{RoyalBlue}FT-dinosaur} &50.3 &21.1&50.1\\    
     {\color{BrickRed}FT-dinosaur} & 51.1&22.6&50.9 \\
  \bottomrule
  \end{tabular}
\end{table}

\subsection{Why Not Segment Anything Model?}
\begin{figure}
    \centering
    \includegraphics[width=0.6\linewidth]{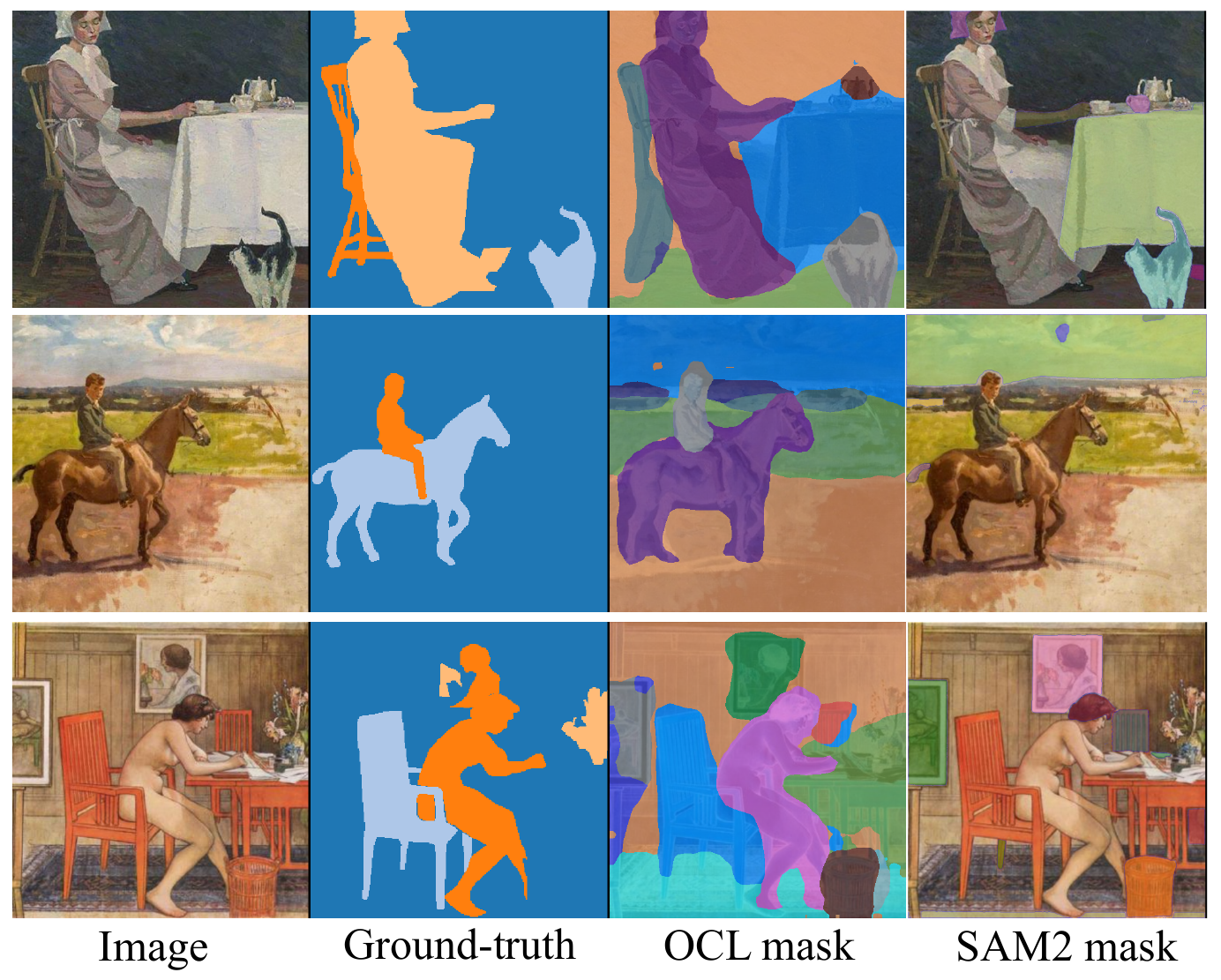}
    \caption{Comparison of SAM2 automatic segmentation and our OCL (FT-dinosaur + learned slot initialization) segmentation. Unlike SAM2, which in automatic mode detects only a sparse set of objects and leaves large portions of the image unsegmented, OCL produces a complete scene decomposition by partitioning the entire image into a fixed number of semantically meaningful regions. Examples are drawn from the DRAM dataset.}
    \label{fig:dram}
\end{figure}
A natural question is whether a supervised segmentation model such as the Segment Anything Model (SAM) family \cite{kirillov2023segment, ravi2024sam2, carion2025sam} could be used in place of OCL for region-level decomposition. 

\begin{figure}
    \centering
    \includegraphics[width=0.6\linewidth]{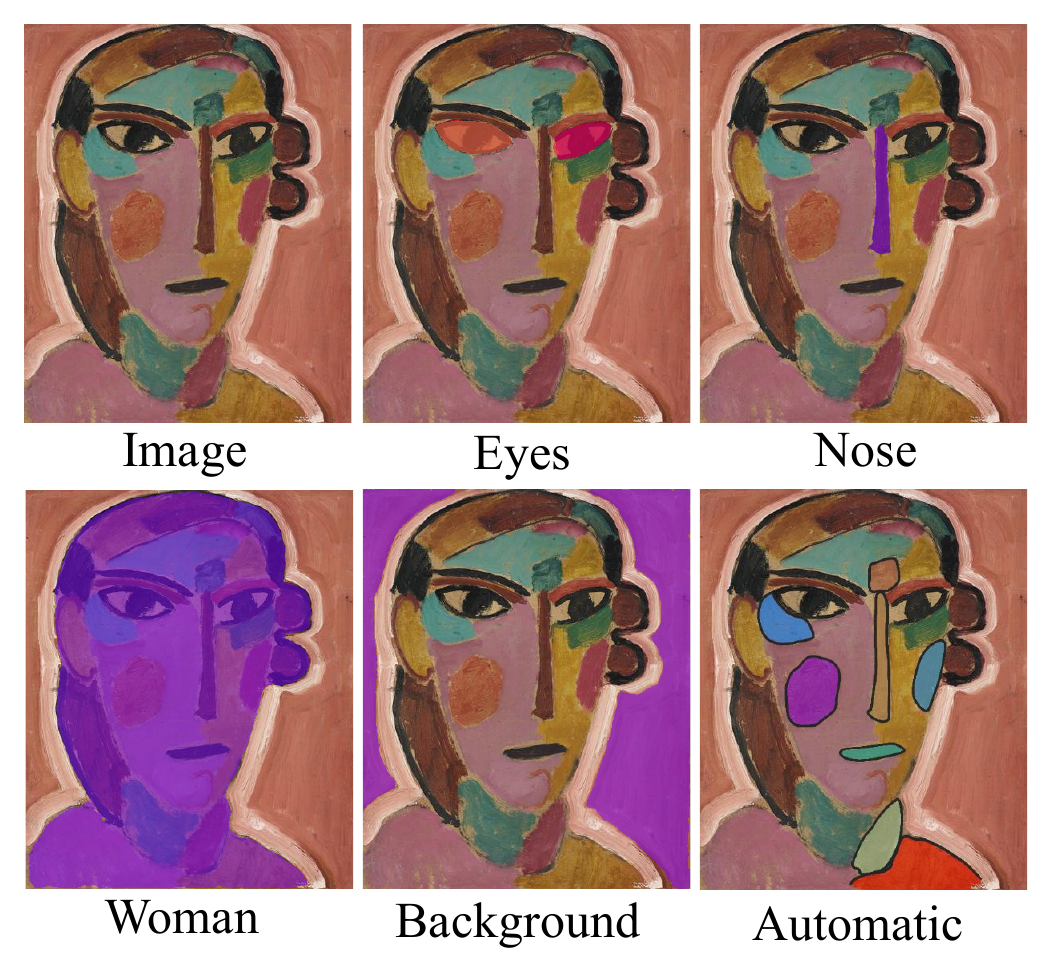}
    \caption{SAM3 \cite{carion2025sam} outputs using different prompts vs. the automatic mode, where no prompt is provided. Having a prompt is a necessary component for the good performance of SAM3. The image is drawn from the DRAM dataset.}
    \label{fig:sam_output}
\end{figure}

We use OCL for two reasons.
First, our goal is not high-quality segmentation, but rather obtaining a compact and meaningful representation per region that can serve as a node in our graph. OCL models are designed precisely for this purpose. 
They produce a fixed set of region-level representations that encode each region, which can then be directly used to learn inter-region relationships. SAM, by contrast, is a segmentation model that produces masks without associated feature representations and is therefore not directly compatible with our relational learning framework.
Second, when used in automatic mode, without any user-provided prompts, SAM does not produce a complete and coherent decomposition of the scene. As illustrated in \cref{fig:dram}, automatic SAM tends to detect a sparse set of salient objects rather than providing full scene coverage, leaving large portions of the image unaccounted for. This is in contrast to OCL models, which partition the entire image into a fixed number of slots by design. We note that SAM can produce high-quality, detailed masks when guided by appropriate prompts (\cref{fig:sam_output}); however, prompt-based segmentation is not applicable in our fully automatic pipeline. For these reasons, OCL models remain the most suitable choice for our framework.

A further limitation is that, unlike OCL, which produces a fixed number of region-level representations by design, SAM yields a variable number of segments per image. This inconsistency is incompatible with our graph-based framework, which requires a fixed number of nodes.

\begin{figure}
    \centering
    \includegraphics[width=0.6\linewidth]{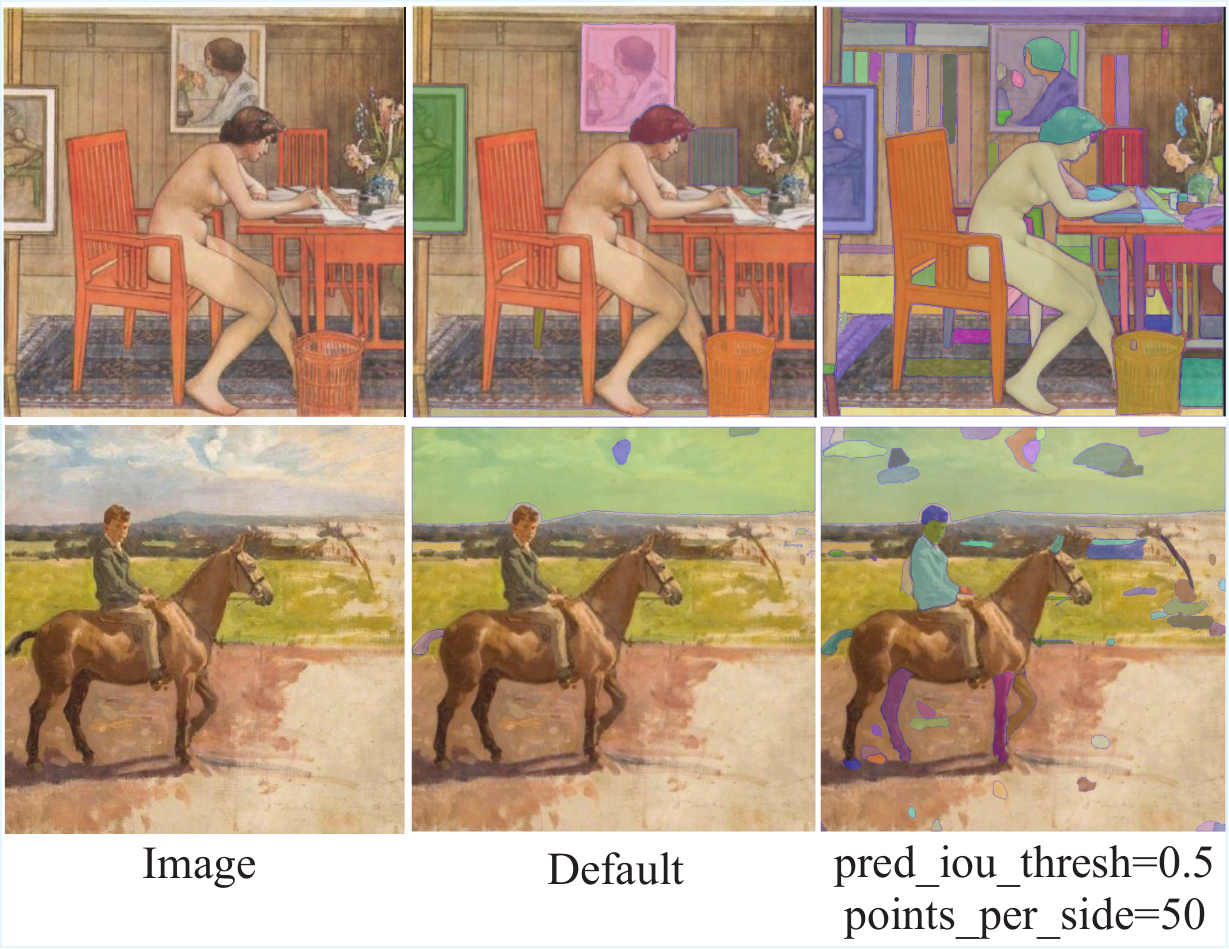}
    \caption{Effect of varying SAM2 parameters on the segmentation output. As shown, optimal segmentation quality requires separate parameter tuning per image, making SAM impractical for use in a fully automatic pipeline.}
    \label{fig:sam_params}
\end{figure}

The SAM family offers several tunable parameters that can improve segmentation quality. \cref{fig:sam_params} shows an example of the effect of changing the parameters. However, this introduces an additional challenge:  while increasing the level of detail in SAM's output can produce finer segmentation masks, such granularity is counterproductive for composition understanding. Learning the global spatial layout of a scene does not require fine-grained object boundaries; rather, it requires a compact, high-level decomposition of the scene into its principal regions. OCL naturally provides this level of abstraction, making it better suited to our task.

\section{GAT Graph on the APDDv2 Dataset}
\begin{figure}
    \centering
    \includegraphics[width=1\linewidth]{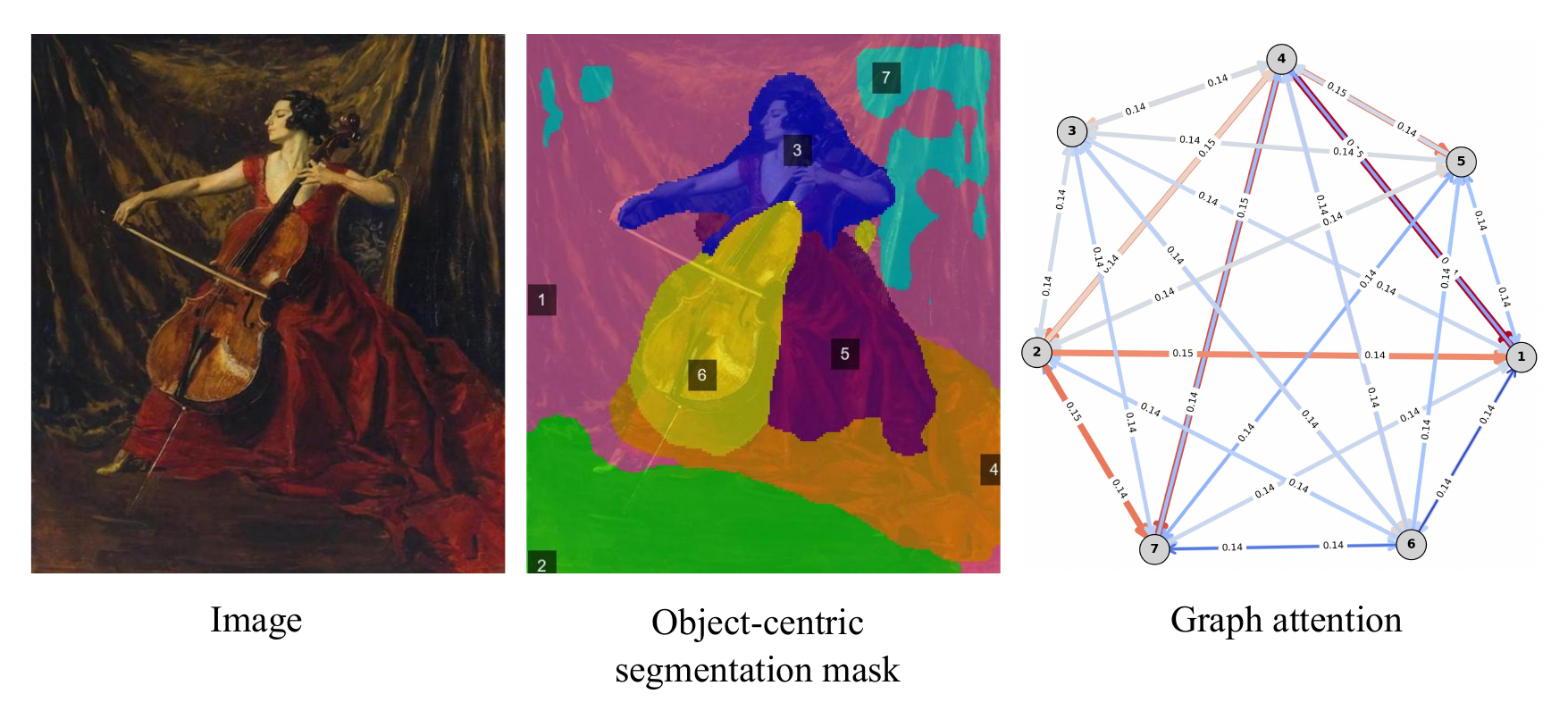}
    \caption{An example of a graph learned on the APDDv2 dataset.}
    \label{fig:apdd_problem}
\end{figure}

\cref{fig:apdd_problem} illustrates the GAT graphs learned on the APDDv2 dataset. As shown, the edge weights are nearly uniform across all nodes. This can be explained by the fact that predicting a global composition score does not fully rely on identifying specific inter-region relationships, and as a result, the GAT does not learn meaningful or discriminative edge weights.

\section{Image Retrieval}
\begin{figure}
    \centering
    \includegraphics[width=1\linewidth]{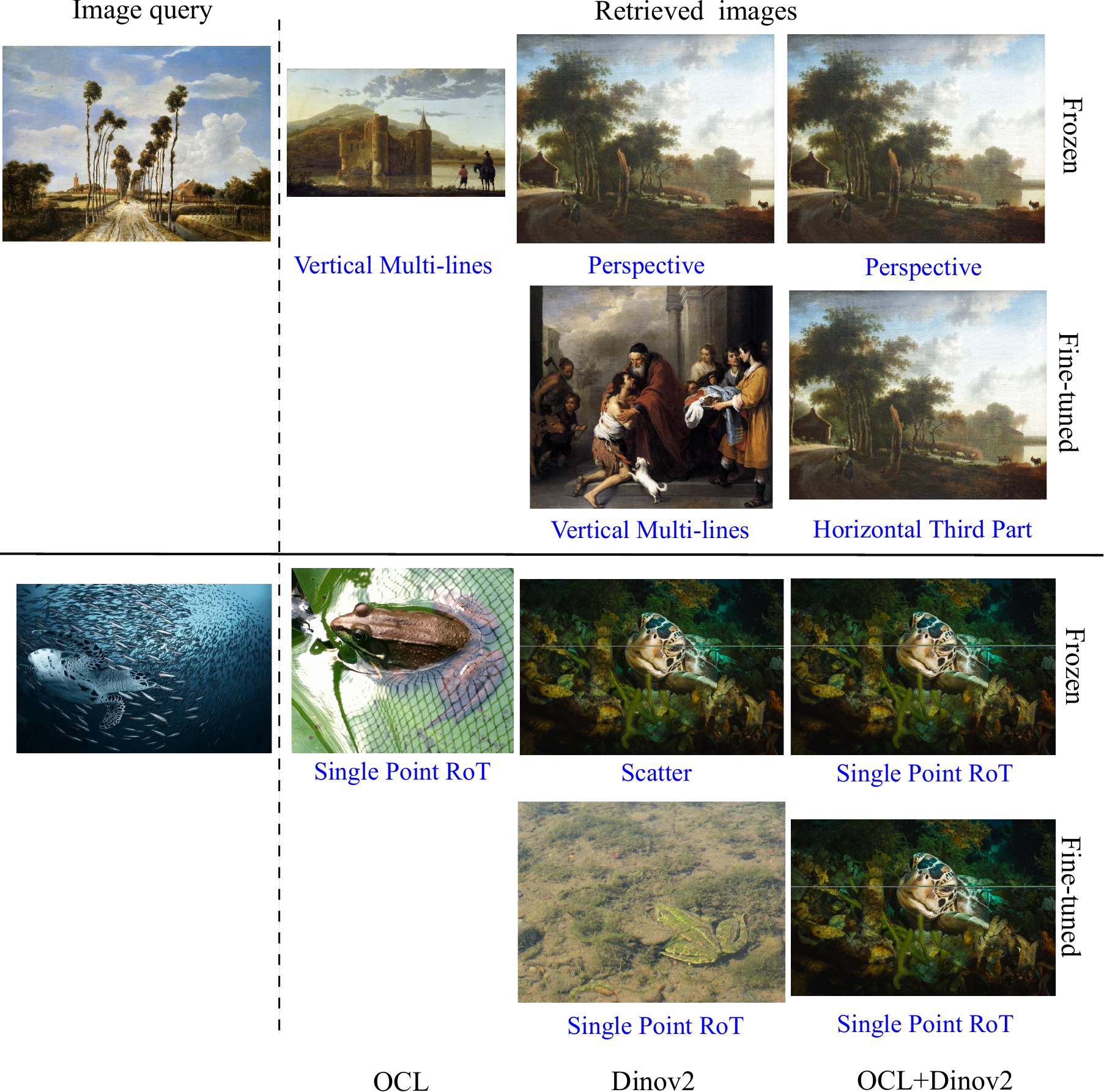}
    \caption{This figure shows the query image and the top retrieved images from each model, along with each model’s prediction. The results indicate that retrieval is strongly driven by the predicted class.}
    \label{fig:image_retrival_supp}
\end{figure}
\cref{fig:image_retrival_supp} shows another example of compositional image retrieval. We observe that 1) retrieval is strongly driven by the predicted composition class, and 2) when a query image belongs to multiple composition categories, the retrieved images tend to share at least one of those categories, though not necessarily the full combination. For example, the first query image in \cref{fig:image_retrival_supp} belongs to three categories of perspective, vertical multilines, and horizontal third, while the retrieved images each belong to at least one of these categories rather than the same combination.

We do not use models trained on APDDv2 for compositional image retrieval for two reasons. First, composition scores are inherently continuous and subject to intra-rater variability — the same image may receive slightly different scores even from the same participant across sessions. Retrieval based on score similarity, therefore, requires defining an arbitrary margin, making evaluation sensitive to this choice. Second, and more fundamentally, a global composition score lacks visual interpretability: two images with identical scores may be compositionally dissimilar in appearance, making retrieved results difficult to validate qualitatively.
For these reasons, composition category-based retrieval provides a more reliable and interpretable evaluation framework.

\section{Visual examples on photographs}

\begin{figure}
    \centering
    \includegraphics[width=1\linewidth]{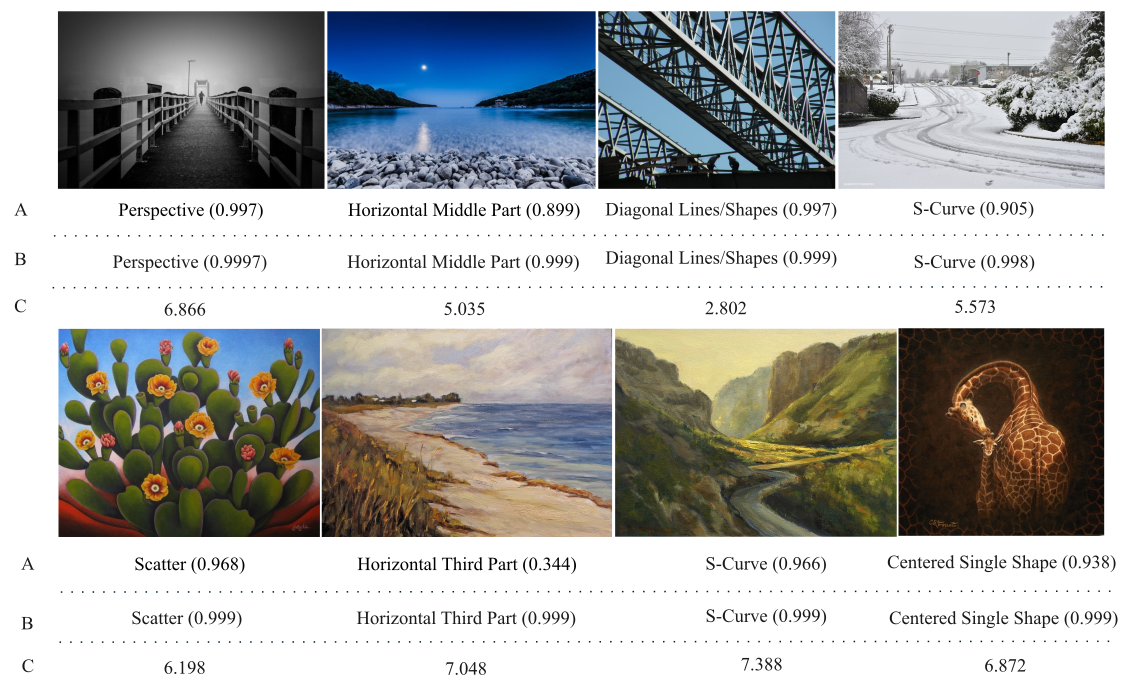}
    \caption{Qualitative examples of composition category and score prediction. Model A, B, and C refer to {\color{RoyalBlue}OCL} + {\color{BrickRed}GAT} + {\color{RoyalBlue}Dinov2-b} fine-tuned on PICD, {\color{BrickRed}Dinov2-b} fine-tuned on PICD, and {\color{RoyalBlue}OCL} + {\color{BrickRed}GAT} + {\color{BrickRed}Dinov2-b} fine-tuned on APDDv2, respectively. Numbers in parentheses indicate the model's confidence score for the predicted category.}
    \label{fig:prediction_example}
\end{figure}

In this section, we provide additional qualitative examples of our best-performing models.

\begin{figure}[t]
    \centering
    \includegraphics[width=0.85\linewidth]{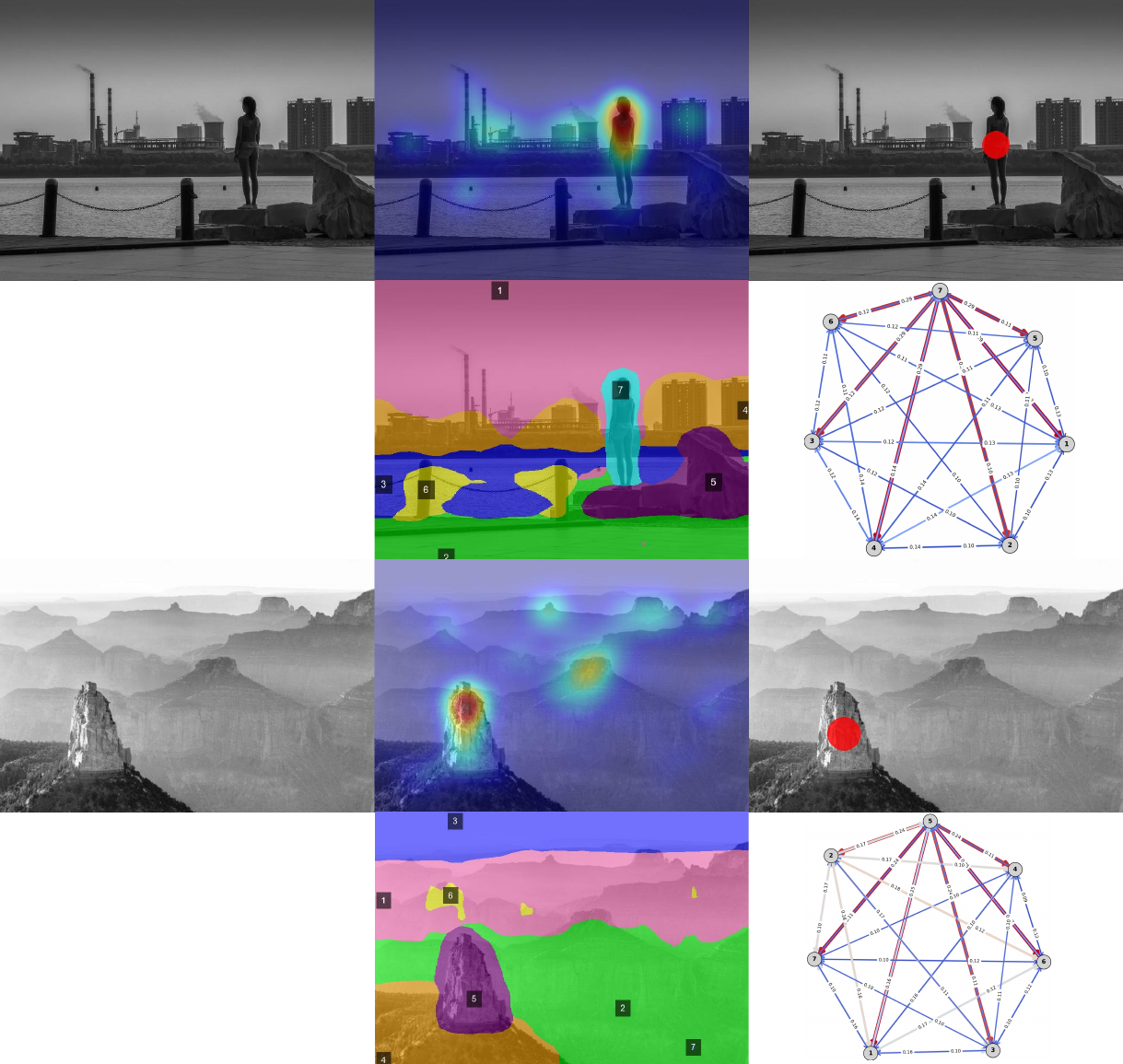}
    \caption{Qualitative examples of visual saliency detection on photographs. For each image, we show: (1) the eye-tracking fixation map from \cite{wagemans2026saccade}, (2) our predicted salient regions, (3) the OCL segmentation mask, and (4) the learned GAT graph.}
    \label{fig:photograph_example}
\end{figure}
\subsection{Composition category and score prediction}
On the PICD dataset, we compare the best frozen setting ({\color{RoyalBlue}OCL} + {\color{BrickRed}GAT} + {\color{RoyalBlue}Dinov2-b}) against the best fine-tuned setting ({\color{BrickRed}Dinov2-b}). As shown in \cref{fig:prediction_example}, fine-tuned {\color{BrickRed}Dinov2-b} not only outperforms {\color{RoyalBlue}OCL} + {\color{BrickRed}GAT} + {\color{RoyalBlue}Dinov2-b} in terms of prediction accuracy, but also produces higher confidence scores for the predicted categories. Interestingly, the composition categories learned on PICD appear to generalize to artwork, suggesting that the compositional concepts captured by the model transfer across domains.

On the other hand, composition scores predicted by {\color{RoyalBlue}OCL-BAID} + {\color{BrickRed}GAT} + {\color{BrickRed}Dinov2-b} fine-tuned on APDDv2 do not generalize to photographs, producing scores that are difficult to interpret in that context. Unfortunately, since APDDv2 does not provide composition category labels and PICD does not provide composition scores, a quantitative evaluation of zero-shot cross-domain performance is not possible with the currently available data.

\subsection{Visual saliency prediction} 
\begin{figure}[t]
    \centering
    \includegraphics[width=1\linewidth]{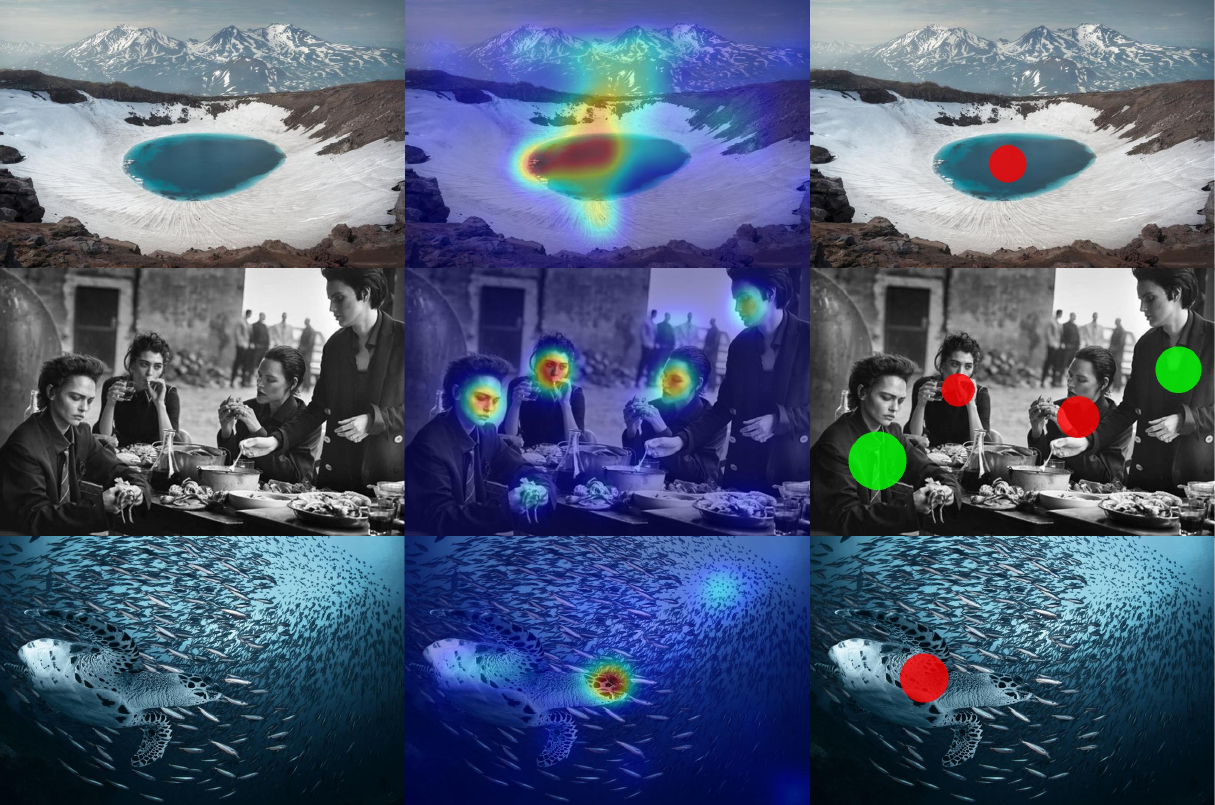}
    \caption{Comparison between visually salient regions detected by our model and human fixation patterns from eye-tracking data on photographs \cite{wagemans2026saccade}. Our model correctly identifies the most salient regions and their relative importance ordering.}
    \label{fig:eye_tracking_supp}
\end{figure}

All graphs and saliency predictions in this paper are produced by our {\color{RoyalBlue}OCL} + {\color{BrickRed}GAT} + {\color{RoyalBlue}Dinov2-b} model trained on PICD. \cref{fig:photograph_example} and \cref{fig:eye_tracking_supp} show examples of visual saliency prediction on photographs.

\section{Attention Faithfulness}

In this study, we use GAT attention weights as a measure of the importance of each image region for compositional prediction. This assumes that the attention weights are \emph{faithful}: regions assigned high attention should be more important for the model's prediction than regions assigned low attention. We test this assumption by measuring how the model's predictions change when regions are removed according to their attention weights~\cite{petsiuk2018rise}.

\textbf{Importance score} For an image with $N=7$ slots, let
$A \in \mathbb{R}^{N \times N}$ denote the GAT attention matrix averaged
over attention heads, where $A_{ij}$ is the attention weight that node $i$
assigns to node $j$. We define the importance of node $i$ as its total
outgoing attention:
\begin{equation}
\text{importance}(i) = \sum_{j=1}^{N} A_{ij}.
\end{equation}

\textbf{Deletion curves} For each test image, we construct three
orderings for progressively removing slots before they are passed to the
GAT and prediction head: (1) \emph{highest-first}, which removes the
highest-importance slot first; (2) \emph{lowest-first}, which removes the
lowest-importance slot first; and (3) \emph{random}, which uses a
uniformly random permutation. For the random condition, we average the
results over three independent permutations to reduce variance.

Given a slot vector $s_i \in \mathbb{R}^D$, removal is implemented by
setting $s_i \leftarrow 0$ before the GAT and MLP prediction head. We do
not resize the slot set, so the graph structure and the number of nodes
remain fixed throughout the experiment. 

At each removal step $k \in {0,\ldots,N}$, we measure the change in the
model's predicted probability relative to the original prediction
obtained before any slots are removed ($k=0$). Because composition
prediction on PICD is a multi-label task, we compute this change only
for classes predicted as positive (probability $>0.5$) in the original
forward pass. We then report the mean probability drop across the test
set as a function of the fraction of removed nodes, $k/N$.

\textbf{Top-$k$ comparison} As a complementary test, we compare model
performance when only the top-3 slots are retained
(all other slots are zeroed) with performance when only the bottom-3
slots are retained. Performance is measured using F1 on the multi-label prediction
task.

\subsection{Results}

As shown in Fig.~7 in the main paper, the three deletion curves exhibit
the expected ordering for faithful attention. Removing the
highest-attention slot first produces the largest initial drop in
prediction confidence. After removing a single node, the mean
probability drop is 0.180, compared with 0.037 for random removal,
corresponding to an approximately $5\times$ larger drop. In contrast,
removing the lowest-attention slot first has little effect on the
prediction until most of the nodes have been removed.

All three curves converge at $k=N$, because all slots are zeroed
regardless of the removal order. The resulting model output for this
all-zero input is not zero: the mean probability drop is approximately
0.40. This is expected because the GAT and MLP contain learned bias
terms that remain active when the slot representations are zeroed.
Therefore, the absolute value at $k=N$ is not itself a measure of
faithfulness. Instead, the relevant evidence is the relative ordering
and separation of the curves during the deletion process.

We summarize the separation between the curves using the area between
the highest-first and random curves, and between the random and
lowest-first curves. These areas are computed by trapezoidal
integration over the fraction of nodes removed. On the full PICD test
set, the resulting areas are 0.1449 and 0.1040, respectively. The
positive separation between the curves is consistent with the
hypothesis that nodes assigned higher attention are more important to
the model's prediction than nodes assigned lower attention.

Table~\ref{tab:attention_faithfullness} reports the top-3/bottom-3 comparison. Keeping only the top-3 slots achieves an F1 score of 73.83\%, compared with 36.87\%
when only the bottom-3 slots are kept. This gap provides complementary
evidence for the deletion-curve analysis: an equally sized subset of
the regions assigned the highest attention is substantially more
predictive than a subset of the regions assigned the lowest attention.

\begin{table}[h]
\centering
\caption{Top-3 vs. bottom-3 on the full PICD test
set.}
\label{tab:attention_faithfullness}
\begin{tabular}{lc}
\hline
Slots kept & F1 (\%) \\
\hline
Top-3 (highest attention) & 73.83 \\
Bottom-3 (lowest attention) & 36.87 \\
\hline
\end{tabular}
\end{table}
 
Overall, these results support the use of GAT attention weights as a
faithful measure of region importance in our model. The regions assigned
higher attention are more important for the model's own predictions,
both when they are progressively removed and when they are evaluated as
a small retained subset. Together with the human saliency analysis in
Section~4.4 (Fig.~6) in the main paper, these results suggest that the GAT identifies
regions that are both visually salient to human observers and
predictively important to the model.

\section{{\color{RoyalBlue}OCL} + {\color{BrickRed}GAT} + {\color{BrickRed}Dinov2-b}}
\begin{figure}
    \centering
    \includegraphics[width=1\linewidth]{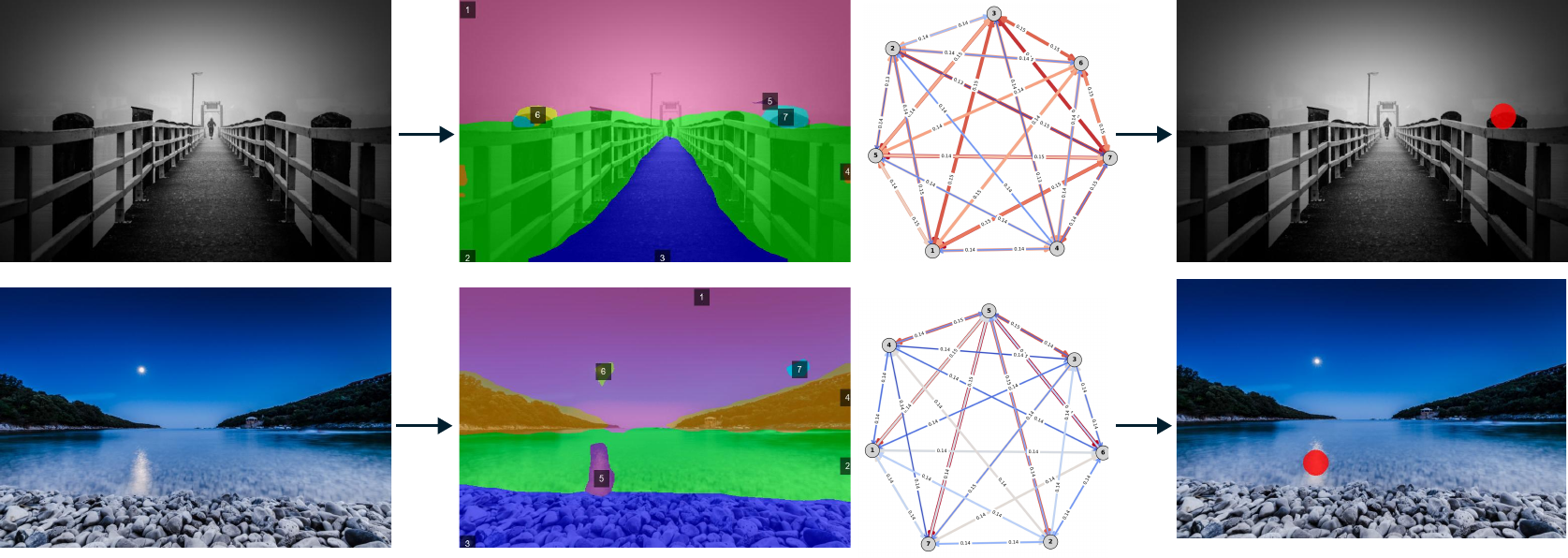}
    \caption{Examples of learned GAT graphs and their corresponding salient region predictions, produced by our {\color{RoyalBlue}OCL} + {\color{BrickRed}GAT} + {\color{BrickRed}Dinov2-B} model.}
    \label{fig:ocl_dinof}
\end{figure}

We also explore a hybrid configuration in which Dinov2-b is fine-tuned while OCL remains frozen. As shown in \cref{fig:ocl_dinof}, this setting produces a near-uniform GAT graph in which all edge weights are almost similar, indicating that the model fails to learn meaningful inter-region relationships. We attribute this to the large imbalance between the two feature sources: Dinov2-b contributes 1,384 tokens compared to only 7 tokens (slots) from OCL, causing the fine-tuned Dinov2-b features to dominate and the GAT to effectively ignore the OCL representations.

The impact of this relational breakdown is shown in \cref{tab:picd_results_ocl_dinof} and \cref{tab:apdd_results_ocl_dinof}. On PICD, where composition is defined by discrete categories strongly tied to spatial arrangements and inter-region relationships, the failure of the GAT to learn meaningful structure means that the OCL branch contributes noise (\cref{fig:ocl_dinof}) rather than a useful signal, resulting in a performance decrease. On APDDv2, however, composition is measured as a continuous global score influenced by broader image properties such as color and contrast, which are well captured by the fine-tuned Dinov2-b features. In this setting, the additional OCL tokens, despite lacking meaningful relational structure, may still provide complementary information that marginally benefits global score prediction, resulting in a slight performance improvement.
\begin{table}
  \caption{Performance of {\color{RoyalBlue}OCL} + {\color{BrickRed}GAT} + {\color{BrickRed}Dinov2-b} on \textbf{PICD test set}. {\color{RoyalBlue}Blue} and {\color{BrickRed}Red} show frozen and fine-tuned modules, respectively.
  }
  \label{tab:picd_results_ocl_dinof}
  \centering
  \begin{tabular}{@{}l|l|l|l@{}}
    \toprule
    Precision (\%) & Recall (\%)& F1-score (\%)& Accuracy (\%)\\
    \midrule
    87.05&86.7&86.88&98.89\\
  \bottomrule
  \end{tabular}
\end{table}

\begin{table}
  \caption{Performance of {\color{RoyalBlue}OCL} + {\color{BrickRed}GAT} + {\color{BrickRed}Dinov2-b} on \textbf{APDDv2 test set}. {\color{RoyalBlue}Blue} and {\color{BrickRed}Red} show frozen and fine-tuned modules, respectively. 
  }
  \label{tab:apdd_results_ocl_dinof}
  \centering
  \begin{tabular}{@{}l|l|l@{}}
  \toprule
  PLCC (\%) & SRCC (\%)& Accuracy(\%) \\
    \midrule
    70.93&69.71&84.07\\
  \bottomrule
  \end{tabular}
\end{table}


\section{Image credits}
\begin{itemize}
    \item Figure 1: Vincent van Gogh, Café Terrace at Night (1888); Arthur Segal, Street with Church Tower (1924); Raphael, Canigiani Holy Family (1507); Dante Gabriel Rossetti, The Bower Meadow (1872).
    \item Figure 2: Gerard van Honthorst, The Matchmaker (1625); Gustave Caillebotte, Paris Street; Rainy Day (1877).
    \item Figure 3: Artemisia Gentileschi, Judith Slaying Holofernes (c. 1620).
    \item Figure 4: Rembrandt van Rijn, Christ in the Storm on the Sea of Galilee (1633). Retrieved images are from the BAID dataset.
    \item Figure 5: Fra Filippo Lippi, Madonna with the Child and Two Angels, c. 1450-65; Ludovico Mazzolino, The Twelve-Year-Old Jesus Teaching in the Temple, 1524.
    \item Figure 6:  Caravaggio, The Supper at Emmaus (1601); Edvard Munch, The Sick Child (1885).
    \item Figure 7: Katsushika Hokusai, The Great Wave off Kanagawa (c. 1831).
    
\end{itemize}


\end{document}